\documentclass[letterpaper]{article} 
\usepackage[draft]{aaai2026}  
\usepackage{times}  
\usepackage{helvet}  
\usepackage{courier}  
\usepackage[hyphens]{url}  
\usepackage{graphicx} 
\usepackage{booktabs}
\usepackage{natbib}  
\usepackage{caption} 
\usepackage{algorithm}
\usepackage{algorithmic}

\usepackage{newfloat}
\usepackage{listings}
\DeclareCaptionStyle{ruled}{labelfont=normalfont,labelsep=colon,strut=off} 
\floatstyle{ruled}
\newfloat{listing}{tb}{lst}{}
\floatname{listing}{Listing}
\title{Studying Cross-cluster Modularity in Neural Networks}

\author {
    Satvik Golechha\textsuperscript{\rm 1},
    Maheep Chaudhary\textsuperscript{\rm 2},
    Joan Velja\textsuperscript{\rm 3},
    Alessandro Abate\textsuperscript{\rm 4},
    Nandi Schoots\textsuperscript{\rm 4}
}
\affiliations {
    \textsuperscript{\rm 1}Machine Learning Alignment \& Theory Scholars (MATS) Program\\
    \textsuperscript{\rm 2}Independent\\
    \textsuperscript{\rm 3}University of Amsterdam\\
    \textsuperscript{\rm 4}Department of Computer Science, University of Oxford\\
    zsatvik@gmail.com, nandi.schoots@kcl.ac.uk
}

\usepackage{microtype}
\usepackage{graphicx}
\usepackage{subcaption}

\usepackage{amsmath}
\usepackage{amssymb}
\usepackage{mathtools}
\usepackage{amsthm}

\usepackage[capitalize,noabbrev]{cleveref}

\theoremstyle{plain}
\newtheorem{theorem}{Theorem}[section]
\newtheorem{proposition}[theorem]{Proposition}

\theoremstyle{definition}
\newtheorem{definition}[theorem]{Definition}

\theoremstyle{remark}

\usepackage{subcaption}
\begin{document}

\maketitle

\begin{abstract}

An approach to improve neural network interpretability is via clusterability, i.e., splitting a model into disjoint clusters that can be studied independently.
We define a measure for clusterability and show that pre-trained models form highly enmeshed clusters via spectral graph clustering. We thus train models to be more modular using a ``clusterability loss'' function that encourages the formation of non-interacting clusters.
We then investigate the emerging properties of these highly clustered models.
We find our trained clustered models do not exhibit more task specialization, but do form smaller circuits. 
We investigate CNNs trained on MNIST and CIFAR, small transformers trained on modular addition, and GPT-2 and Pythia on the Wiki dataset, and Gemma on a Chemistry dataset.
This investigation shows what to expect from clustered models.

\end{abstract}

\section{Introduction}

Interpretability is an active area of research that aims to solve both high-stake deployment constraints for fairness and robustness~\citep{McClure2020EvaluatingAR} and AI safety and trustworthiness concerns~\citep{Bereska2024MechanisticIF}. 
Several breakthroughs in the subdomain of mechanistic interpretability, have helped us understand the inner workings of deep networks, both via circuits~\citep{wang2022interpretability, olah2020zoom, elhage2021mathematical} and representation spaces~\citep{zou2023representation, bricken2023towards}.

One approach to mechanistic interpretability is to identify features at the level of neurons, alternatively we can identify `subskills' at the level of subnetworks or circuits.
While sparse auto-encoders (SAEs) \citep{huben2024sparse} extract a model's features over different neurons, we attempt to split a model's computation into separate modules and interpret them in isolation. However, this is feasible only if the interaction, between these modules is minimal, i.e. if the clusterability is high. We introduce a metric to measure the amount of clusterability in neural network components, and attempt to make models modular and more interpretable during training by optimizing for this metric.

\begin{figure}[h]
\centering
    \includegraphics[width=0.42\textwidth,  trim=40 480 40 30, clip]{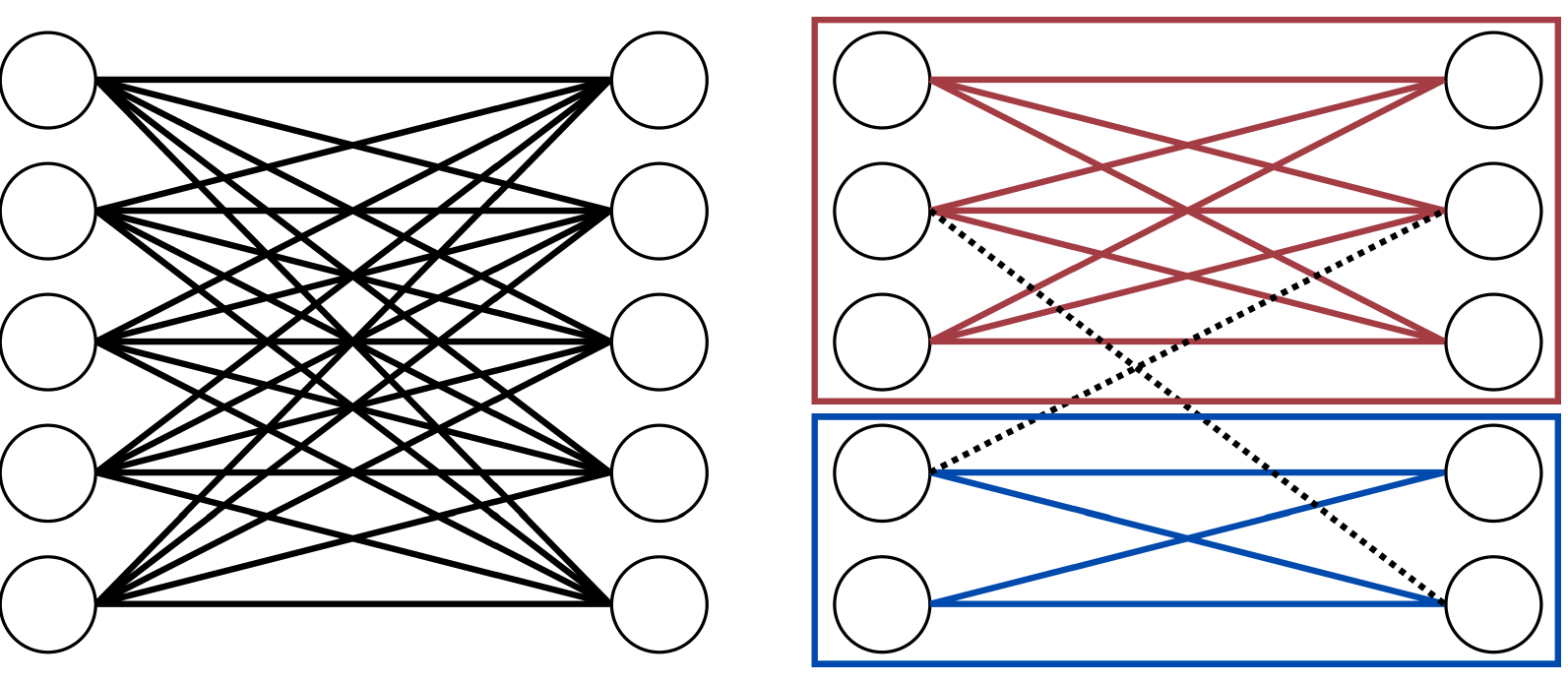}
    \caption{Our training methodology promotes the formation of modular clusters of this kind in any network component.}
    \label{fig:fig1}
\end{figure}

We empirically show that when we train a model to be modular, the modules do \emph{not} specialize any more than non-modular models on the tasks that we test.
For models trained on CIFAR, we find in Figure \ref{fig:classwide-f1} that modules specialize on different labels, but so do non-modular models in Figure \ref{fig:classwide-f2}.
For Gemma trained on a chemistry dataset, we find hardly any module specialization for either modular or non-modular models, see Figure \ref{fig:llm-chemistry}.
For GPT-2 and Pythia we find in Figure \ref{fig:module-comparison} that both modular and non-modular modules, are typically unable to solve a datapoint in isolation.
In other words, we do not find any task specialization of modules.
There may still be task specialization, on a different level of granularity than the ones we investigated.

Applying mechanistic interpretability methods to large models \citep{kaplan2020scaling} and complex behaviors has been a major hurdle for interpretability~\cite{lieberum2023does, golechha2024challenges} due to complicated computational subgraphs (circuits) and superposition~\cite{henighan2023superposition}, i.e., networks representing more features than they have neurons.
We theoretically show in Section \ref{sec:theoretical-representations} that the latent space of a modular matrix has a much smaller feature-space, and empirically show for CIFAR in Section \ref{sec:circuit-size} that splitting models into separate components makes circuits smaller, making pruning and patching \citep{conmy2023towards} more tractable.


The main contributions of our work are as follows:
\begin{itemize}
    \item We introduce a measure of modularity for neural network components and define ``clusterability loss'', a simple and effective regularizer that encourages the emergence of non-interference between clusters during model training (Section \ref{sec:clusterability}). See Figure \ref{fig:fig1} for a visual description of modularity in a layer.
    \item We use automated interpretability methods to show that the clusters thus obtained do not obviously produce specialized clusters  (Section \ref{sec:cluster-specialization}).
    \item We empirically (Section \ref{sec:circuit-size}) and theoretically (Section \ref{sec:theoretical-representations}) show that the clusters reduce the search-space of circuit-style analyses by reducing effective circuit size.
\end{itemize}

We also modify and test prior clusterability methods \cite{filan2021clusterability} to split a trained neural network layer into clusters and show that the clusters thus found are highly enmeshed (not modular) in Section \ref{sec:spectral-clustering}, discuss how our training methodology, as described in Section \ref{subsec:training}, can help with building or fine-tuning more modular models, and discuss some mathematical implications in Section \ref{sec:theoretical}.



\section{Related Work}
\label{sec:related}

\paragraph{Modularity} metrics inspired by research in neuroscience incorporate transfer entropy \citep{novelli2019deriving, ursino2020transfer} 
or spatial metrics \citep{liu2023seeing, liu2023growing}. 
\citet{liu2023growing} incorporate neuron swapping, which is  a discrete search over optimal neuron ordering. This swapping requires $\mathcal{O}(n k L)$ computations, where $n$ is the number of neurons going into a layer, $k<n$ is the number of neurons being sorted, and $L$ is the number of layers.
Since swapping is expensive it is only performed every so many training steps.
Models are sometimes trained or pruned using modularity metrics \citep{patil2023neural}.
\citet{tsitsulin2023graph} find that Graph Neural Networks node pooling methods do not work well to cluster graphs and introduce an unsupervised clustering method.
Their method first uses a graph convolutional network to obtain soft clusters for each node and then 
optimizes this assignment based on the first modularity metric in Section \ref{sec:clusterability}.
\citet{salha2022modularity} add a regularization term for modularity in Graph Autoencoders to show performance gains. In this work, we investigate the relationship between modularity and interpretability.

\paragraph{Mixture of experts} is a form of modularity, where each module is an entire network. In our set-up each module is a component of a layer, i.e. the partitioning is at a lower level. Mixture of experts sometimes use routing policies to decide which expert to activate (e.g. based on features of the input).
Gradient routing \citep{cloud2024gradient} is a method that trains task-specialized experts by hard-coding what tasks each expert specializes on, one of their stated benefits is that this allows a user to selectively remove ability on a particular task. 
MONET \citep{park2024monet} trains monosemantic experts, which learn to route different types of inputs to specific experts during training, without human-specified task assignments.
We investigate the emergence of task-specialized layer components during training.

\paragraph{Clustering} of neural network weights is typically done either using graph properties of the weights (structural)~\citep{watanabe2018modular, filan2021clusterability, patil2023neural} or using correlations between neuron activations (functional) \citep{hod2022detecting, lange2022clustering}.
MoEfication groups feedforward neurons in a pre-trained language model into clusters of `experts' and at inference only activates the most clustered neurons~\citep{zhang2022moefication}. In this paper, we consider both weight-based and gradient-based clustering, and show that they do not help with modularity across clusters.

\paragraph{Circuit Discovery} is an active field of research \citep{wang2022interpretability, olah2020zoom, elhage2021mathematical, conmy2023towards}, which aims to uncover subnetworks that perform a specific functionality. 
\citet{wortsman2020supermasks} use a randomly initialized, fixed base network and for each task find a subnetwork that achieves good performance on this task. 
These overlapping subnetworks can be thought of as circuits.
In this work, we investigate if modularity can help with reducing circuit size.

\section{Introducing a Measure of Modularity}
\label{sec:clusterability}

First, we discuss existing modularity metrics used in other contexts and their drawbacks in our setting. We then motivate our choice of the clusterability metric (and the clusterability loss) and discuss its benefits as an optimization metric for neural network modularity.

Non-differentiable metrics to measure modularity that rely on conditional entropy \citep{ursino2020transfer}, sampling, or discrete computation \cite{veniat2021efficientcontinuallearningmodular} have been introduced. While they capture the essence of modularity, we cannot use gradient descent to optimize for them during training.
For unweighted graphs, `community structure' is a commonly used modularity metric \citep{newman2006modularity, salha2022modularity, tsitsulin2023graph, bhowmick2024neural}, which is based on the difference between an edge value (0 or 1) and the expected number of edges (see Appendix \ref{sec:community-structure} for an explicit formula).

Our metric, inspired by community structure, is \textit{differentiable} and better tailored to weighted graphs. In particular, in our metric large weights are disproportionally punished, and when we slot in real numbers for the weights, the terms can not cancel out. 

We measure the modularity of a model component by choosing a clustering of the component and calculating the amount of ``clusterability'' in the clusters obtained, i.e., the average fraction of weights that are inside a cluster as opposed to between clusters.


Let \( W \) be the weight matrix where \( W_{ij} \) represents the edge weight between nodes \( i \) and \( j \) in the layer's input and output neurons respectively. Define \( U \) and \( V \) as the clusters for the rows and columns of \( W \), respectively. Let \( C_U(u) \) and \( C_V(v) \) denote the sets of nodes in clusters \( u \in \{1, \ldots, k\} \) and \( v \in \{1, \ldots, k\} \), respectively.
The clusterability measure \( C \) is defined as follows:

\begin{align*}
&C = \frac{\sum\limits_{i=1}^{n} \sum\limits_{j=1}^{n} W_{ij}^2 \cdot \mathbb{I}_{(i \in C_U(u) \land j \in C_V(v))}}{\sum\limits_{i=1}^{n} \sum\limits_{j=1}^{n} W_{ij}^2}, \\
&\text{ where }
\mathbb{I} =
\begin{cases} 
1 & \text{if } i \in C_U(u) \text{ and } j \in C_V(v) , \\ 
  & \text{i.e. if $i$ and $j$ are in the same module} \\
0 & \text{otherwise}
\end{cases}
\end{align*}


We use this metric to measure modularity in neural network components and train models to be more modular by jointly maximizing for clusterability by adding the clusterability loss to the usual cross-entropy loss:
\[\mathcal{L} = \mathcal{L}_{CE} - \lambda \mathcal{L}_C,\]
for an appropriate clusterability coefficient $\lambda$.

Optimizing for this modularity metric has several benefits:


\begin{itemize}
    \item Clusterability measures the extent to which modules are disjoint components which can be studied in isolation.
    \item The metric is differentiable, which makes it possible to optimize for it using back-propagation across a model's parameters.
    \item It is easy to compute. For a model with $k$ components of dimension $n \times n$ that are trained for modularity, computing the clusterability loss takes $O(n^2)$ time and constant space.
    \item 
    In this paper we apply the modularity metric to a single weight matrix, but we can easily generalize the approach to multiple weight matrices. To create modules that are multiple layers deep, one should ensure the neurons in a single cluster in the output layer of the first matrix match the neurons in the input layer of the second matrix and so on.
    \item Mathematically, as discussed in Section \ref{sec:theoretical}, optimizing for our metric leads to models learning simpler function spaces (fewer polytopes) and fewer features (orthogonal directions) to perform a given task, which can be computationally beneficial when model interpretability or explainability is important.
    \item As we show in Section \ref{sec:circuit-size}, optimizing for clusterability leads to smaller circuits (in the case of CIFAR-10).
\end{itemize}

\section{Neural Networks trained on Cross-Entropy are not Modular}
\label{sec:spectral-clustering}

First, we would like to evaluate (based on our clusterability metric) how modular neural networks trained using the cross-entropy loss are by default. Directly searching for clusters that maximize our metric leads to a combinatorial explosion, so we use a clustering algorithm to split a component into clusters. For our clustering algorithm, we modify the methodology of \citet{filan2021clusterability} and use normalized spectral clustering to split any component of a neural network into $k$ different clusters, taking into account that a network layer is bipartite.

Our clustering method, called Bipartite Spectral Graph Clustering (BSGC) is shown in Appendix \ref{app:algorithm} Algorithm \ref{alg:spectral_clustering}. The similarity matrix for BSGC can be created by either the weights of the model or the accumulated gradients.

\paragraph{Weight-based BSGC.}
\label{subsec:weight-bsgc}

Here, we use the weight matrix of a layer as the similarity matrix between neurons of adjacent hidden layers, based on the idea that neurons with strong weights connecting them can be expected to cluster well.

\paragraph{Gradient-based BSGC.}
\label{subsec:grads-bsgc}

Analyzing the gradients of each parameter during training gives us another way to cluster models. The idea is that weights that update together are likely to be part of the same circuit and connect neurons that cluster well together. We set the similarity matrix in Algorithm~\ref{alg:spectral_clustering} to the average cosine similarity of the gradients of each parameter.

\begin{figure}[h]
\vspace{-0.2cm}
\centering
\includegraphics[width=0.45\textwidth,  trim=2 2 40 30, clip]{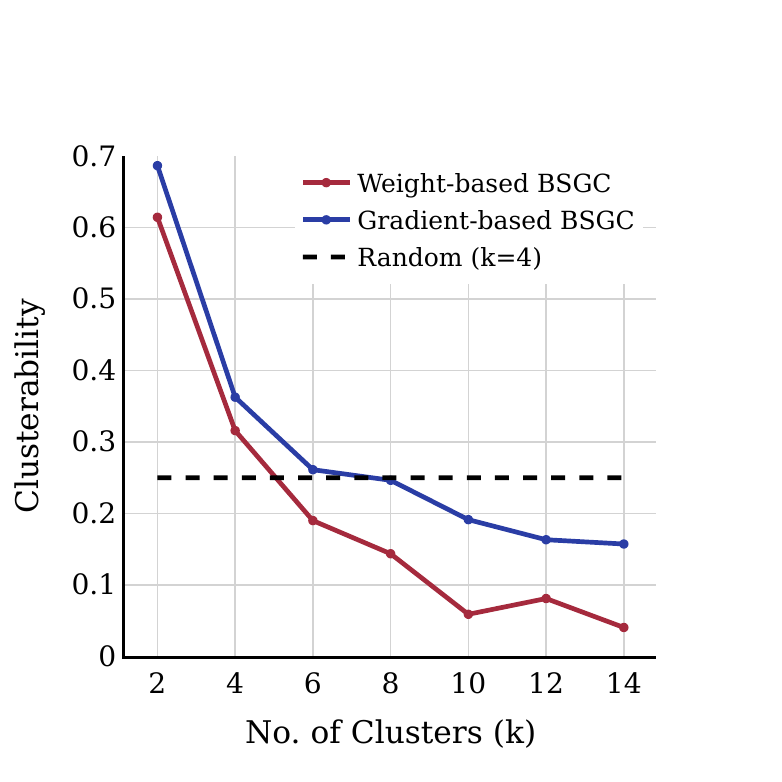}
\vspace{-0.1cm}
    \caption{Clusterability of the model with $k$ clusters using Algorithm~\ref{alg:spectral_clustering} on CIFAR-10. Note that while gradient-based BSGC helps, the resulting clusters are still highly enmeshed and not far above random clusterability, which is $\frac{1}{k}$.}
    \label{fig:clusterability}
\end{figure}

In Figure \ref{fig:clusterability}, we see that as the number of clusters increases, the amount of clusterability decreases, and even at $k=2$ clusters, we get $C=0.6$ (random is 0.5, higher is better), which is substantial interference between clusters. 
We show similar results for Pythia-70m \cite{biderman2023pythiasuiteanalyzinglarge} in Appendix \ref{app:pythia}.

In the remainder of this paper, we arbitrarily choose $k=4$. In general, we would recommend choosing $k$ based on task specific considerations, possibly doing a sweep.

\section{Training for Modularity}
\label{subsec:training}

We train models to be modular by following a training pipeline that comprises the following steps:

\begin{enumerate}
    \item Train the original model to minimize cross-entropy for the first $t$ steps. This can help to form simpler features before we begin clustering and allow ``winning tickets'' (as defined in the lottery ticket hypothesis \citep{Frankle2018TheLT}) to emerge. In practice, we found that our results do not vary with $t$, and default to $t=0$ for most results.
    \item For the set $U$ of model components to cluster, use a clustering method to cluster them and calculate the clusterability loss for each component. Here, we find that gradient-based BSGC (see Algorithm \ref{alg:spectral_clustering}) leads to slightly better initializations (see Figure \ref{fig:clusterability}), but the gains do not remain after our training, which is why we use arbitrary clustering instead. Without loss of generality, we make contiguous clusters of the same size for our experiments for better visualization.
    \item Calculate the effective loss function $\mathcal{L}_{\text{eff}} = \mathcal{L}_{\text{CE}} + \lambda \displaystyle\sum_{u \in \mathcal{U}} \mathcal{L}_{\text{C}}^u$, where $\lambda$ is a hyperparameter that controls the trade-off between performance and modularity. We share results for training on $\lambda=40$, but have found results to be stable across various values of $\lambda$.
    \item Complete the rest of the training for $t$ steps to minimize $\mathcal{L_{\text{eff}}}$ to promote the clusters to be modular (see Section \ref{sec:clusterability} for more details).
\end{enumerate}

\begin{figure*}[t]
\centering
\begin{minipage}[t]{0.48\textwidth}
    \centering
    \includegraphics[height=5cm, trim=2 2 0 50, clip]{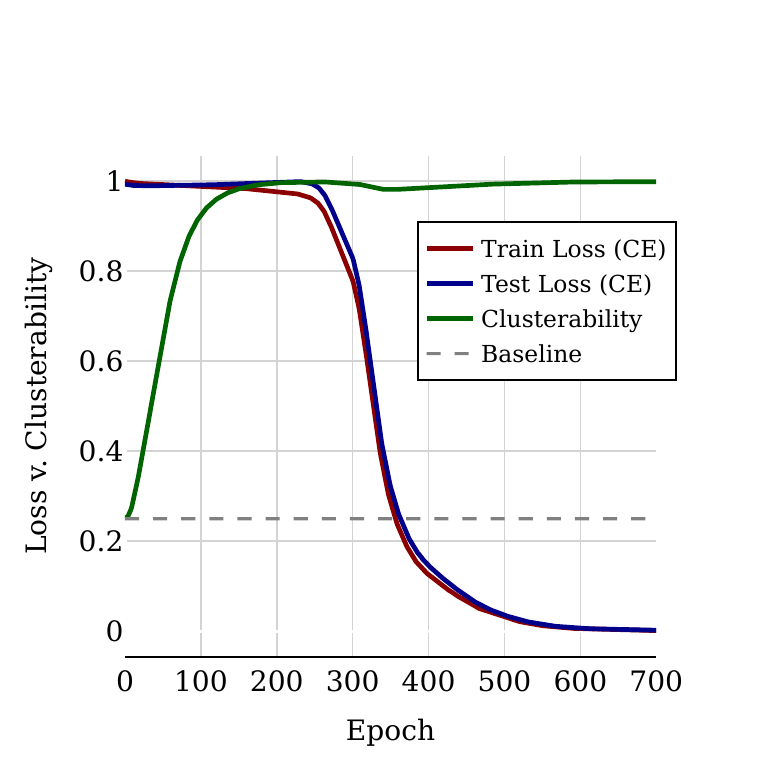}
    \caption{Clusterability vs. train and test losses of a 1-layer transformer trained modularly on modular addition.}
    \label{fig:mod-mod-training}
\end{minipage}
\hfill
\begin{minipage}[t]{0.48\textwidth}
    \centering
    \includegraphics[height=5cm, trim=10 50 40 10, clip]{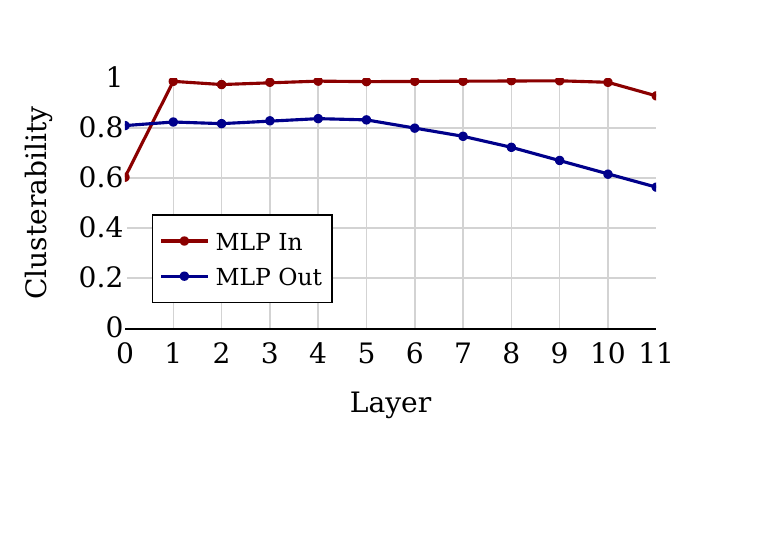}
    \caption{Maximum clusterability of MLP input/output matrices in GPT2-small without degrading performance.}
    \label{fig:mlp}
\end{minipage}
\end{figure*}


We experiment with vanilla MLPs on MNIST \citep{deng2012mnist}, and CNNs on the CIFAR-10 dataset \citep{krizhevsky2009learning}, transformers on modular addition, and fine-tuning of large language models (LLMs). We compare the performance, clusterability, and interpretability gains of our clustered models against models trained without the clusterability loss. Our results are available in Section \ref{sec:results} and all the hyperparameters we use are given in Appendix \ref{app:hyper}.

For modular addition, we follow the training pipeline of \citet{nanda2023progressmeasuresgrokkingmechanistic} to train a 1-layer transformer model and increase the training data to reduce ``grokking'', or the delayed generalization effect they studied. Instead, we focus on training the model with the clusterability loss, and show in Figure \ref{fig:mod-mod-training} that we can train the model to completely solve the task while making every model component completely clusterable. See Appendix \ref{app:arithmetic} for an exploration of the interpretability gains from modularity in this setting.

For language models, we fine-tune GPT-2 \citep{lee2019patentclaimgenerationfinetuning} and Pythia \citep{biderman2023pythia} on a subset of the Wikipedia data \cite{merity2016pointer}, while making the MLP input and output matrices to be modular. 
We also fine-tune
Gemma \citep{gemmateam2024gemma2improvingopen} on a Chemistry dataset \citep{XythicKChemistry}.
For individual layers, we define the ``maximum clusterability'' of a layer to be the maximum amount of clusterability that can be achieved without loss of performance. We measure this by training the model with the clusterability loss until the clusterability stops improving or the performance starts degrading. We show the maximum clusterability for the MLP input and output matrices of each layer in Figure \ref{fig:mlp}.

We show the training plots for modular training of all MLP\_in weight matrices with the cross-entropy training loss in Figure \ref{fig:wiki-training}. We checkpoint the model after the first epoch to avoid overfitting and study the model for interpretability gains, comparing it with a model fine-tuned on the same data for the same number of epochs but without the clusterability loss. The jumps in the cross-entropy loss are due to the model learning from a small enough batch to affect performance on the other datapoints.

\section{Properties of Modular Models}
\label{sec:results}

\subsection{Task-Specialization of Clusters in a Network Trained to be Modular}\label{sec:cluster-specialization}

In this section we do two investigations into the extent to which the modules of a modular model specialize for specific tasks.
We test task-specialization for CNNs on CIFAR in Section \ref{sec:cifar} and find that both modular and non-modular models specialize into specific classes.
In contrast, in Section \ref{sec:gemma-on-chem} we also test whether modules specialize to specific categories of chemistry problems, and find that neither modular nor non-modular models do.
We hypothesize that whether modules specialize or not is related to whether features are localized or not. 

\subsubsection{Modules Specialize in Class-level Features for Both Modular and Non-Modular Models}\label{sec:cifar}

\begin{figure*}[t]
  \centering
  \subfloat[Clustered]{\includegraphics[width=0.48\linewidth,  trim=2 2 50 25, clip]{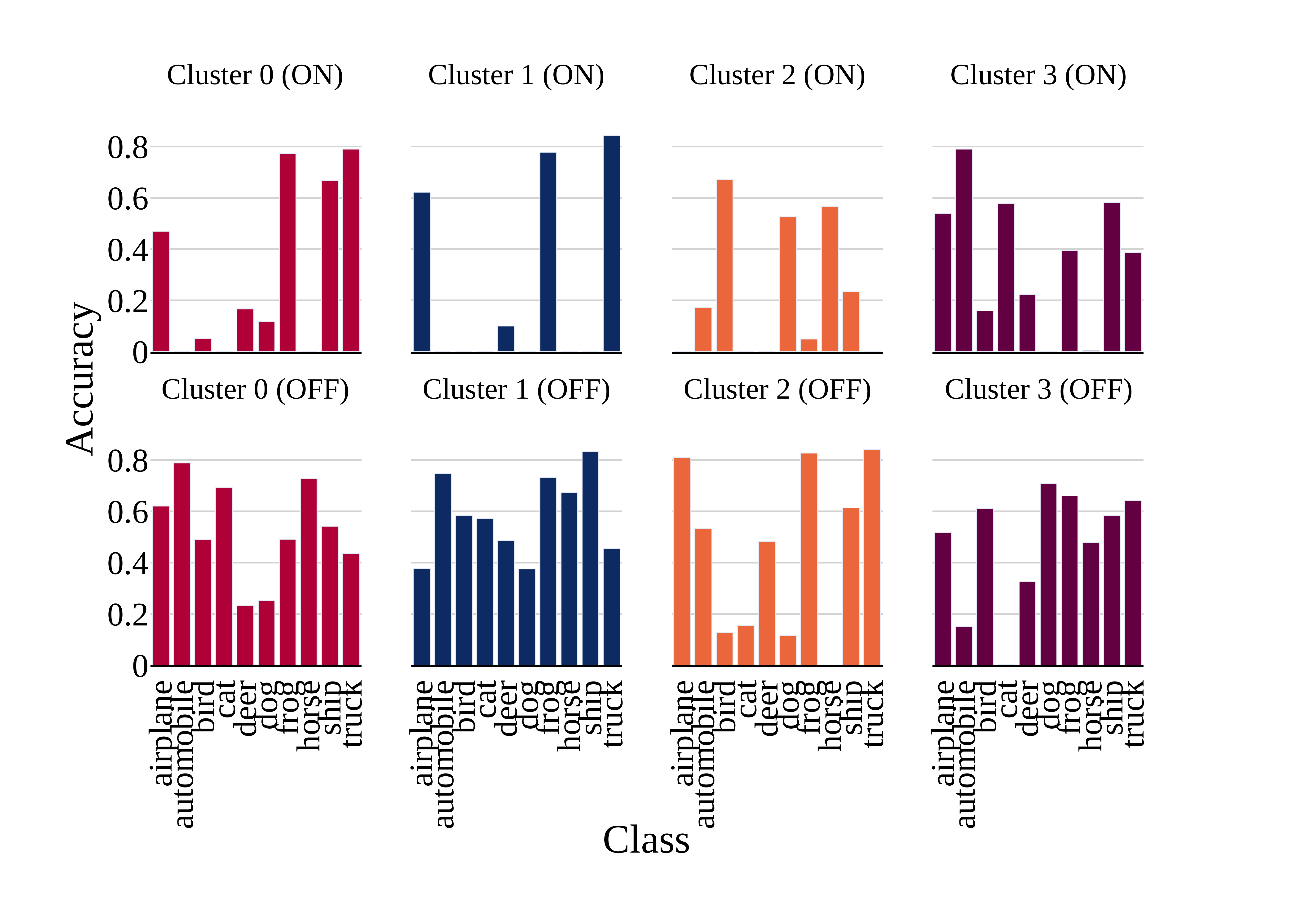}
  \label{fig:classwide-f1}}
  \hspace{0.02\linewidth}
  \subfloat[Unclustered]{\includegraphics[width=0.45\linewidth,  trim=2 2 50 25, clip]{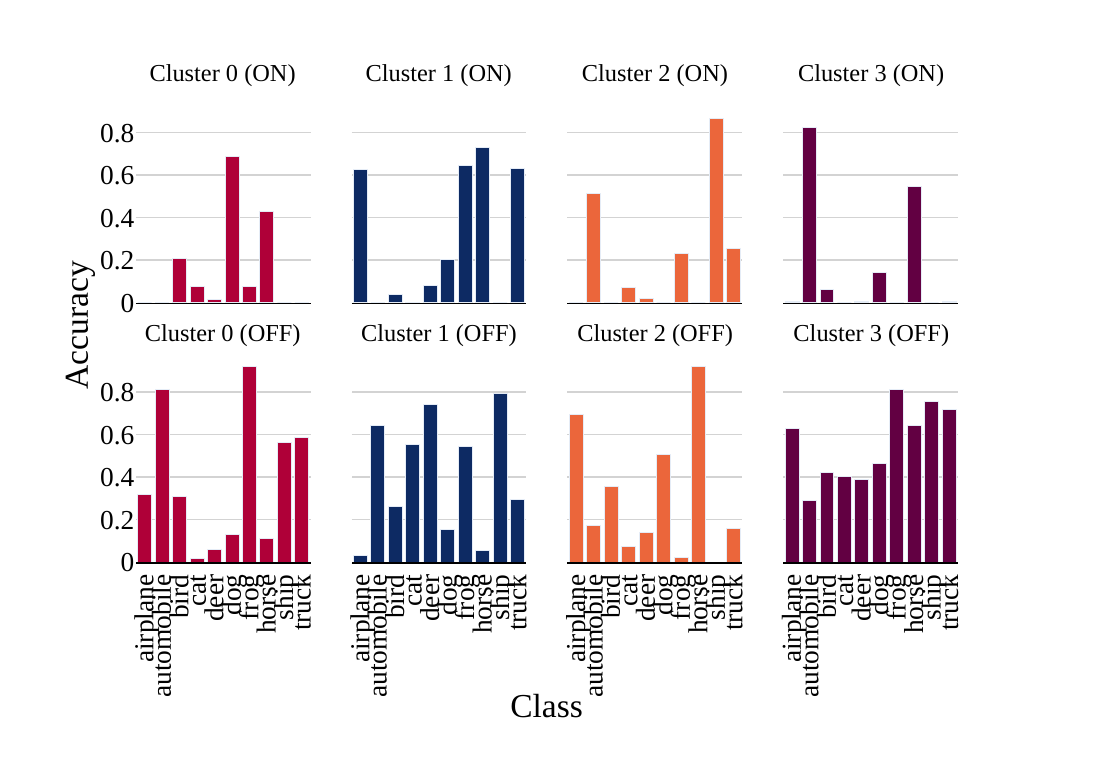}
  \label{fig:classwide-f2}}
   \caption{Class-wise accuracy for each label with clusters turned ON (top) and OFF (bottom). Note that individual clusters learn near-complete circuits for various labels, such as Cluster $0$ in the clustered model for \textit{SHIP} and Cluster $1$ for \textit{TRUCK} for the modular model. 
   The figure on the top shows that a cluster is ``sufficient'' to predict a given class, while a dip in the accuracy in figure on the bottom shows that it is ``necessary''. 
   }
  \label{fig:classwide-accuracies-both}
\end{figure*}

\begin{figure*}[t]
  \centering
  \subfloat[Modular]{\includegraphics[width=0.4\linewidth,  trim=2 40 50 75, clip]{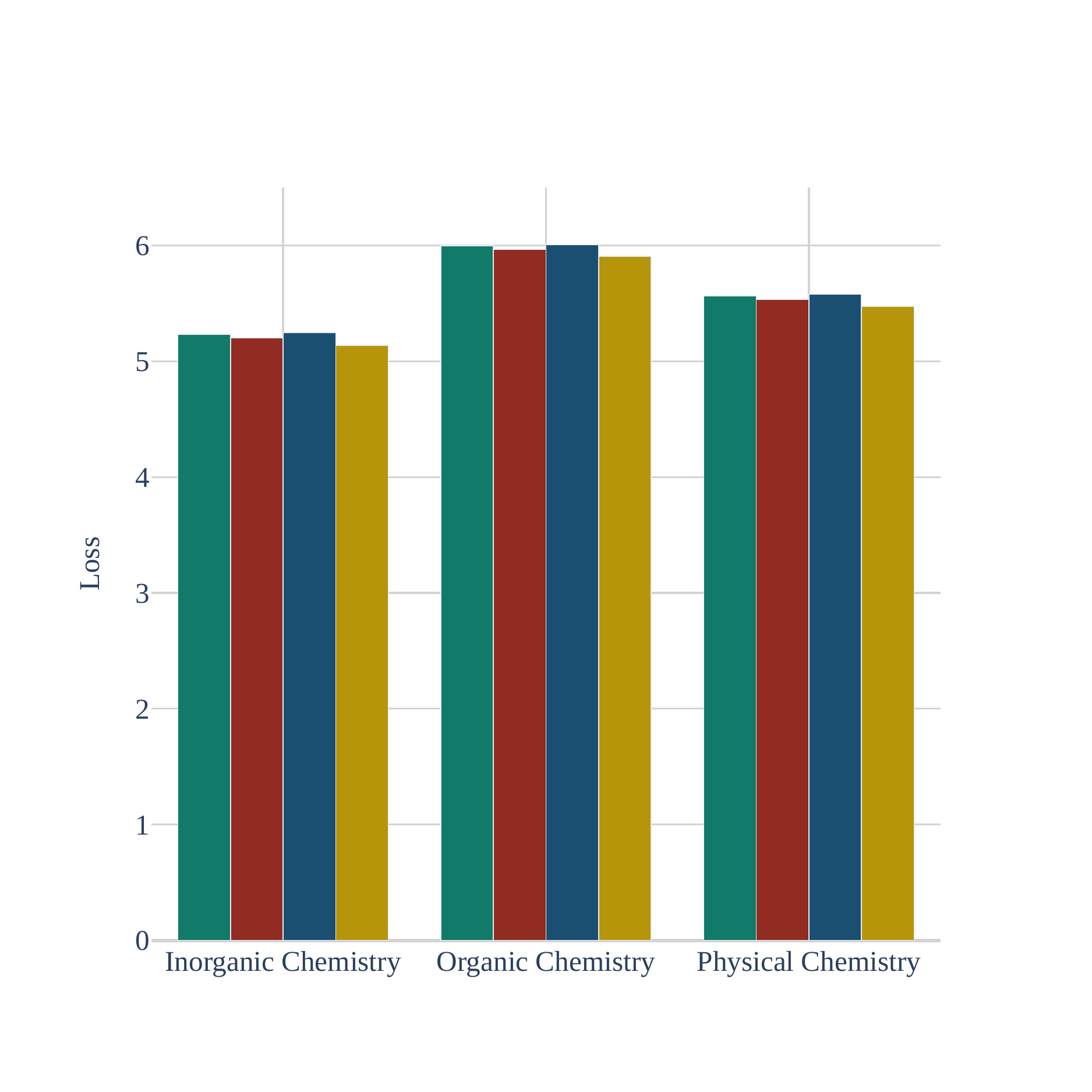}
  \label{fig:classwide-f1}}
  \hspace{0.02\linewidth}
  \subfloat[Non-modular]{\includegraphics[width=0.4\linewidth,  trim=2 40 50 75, clip]{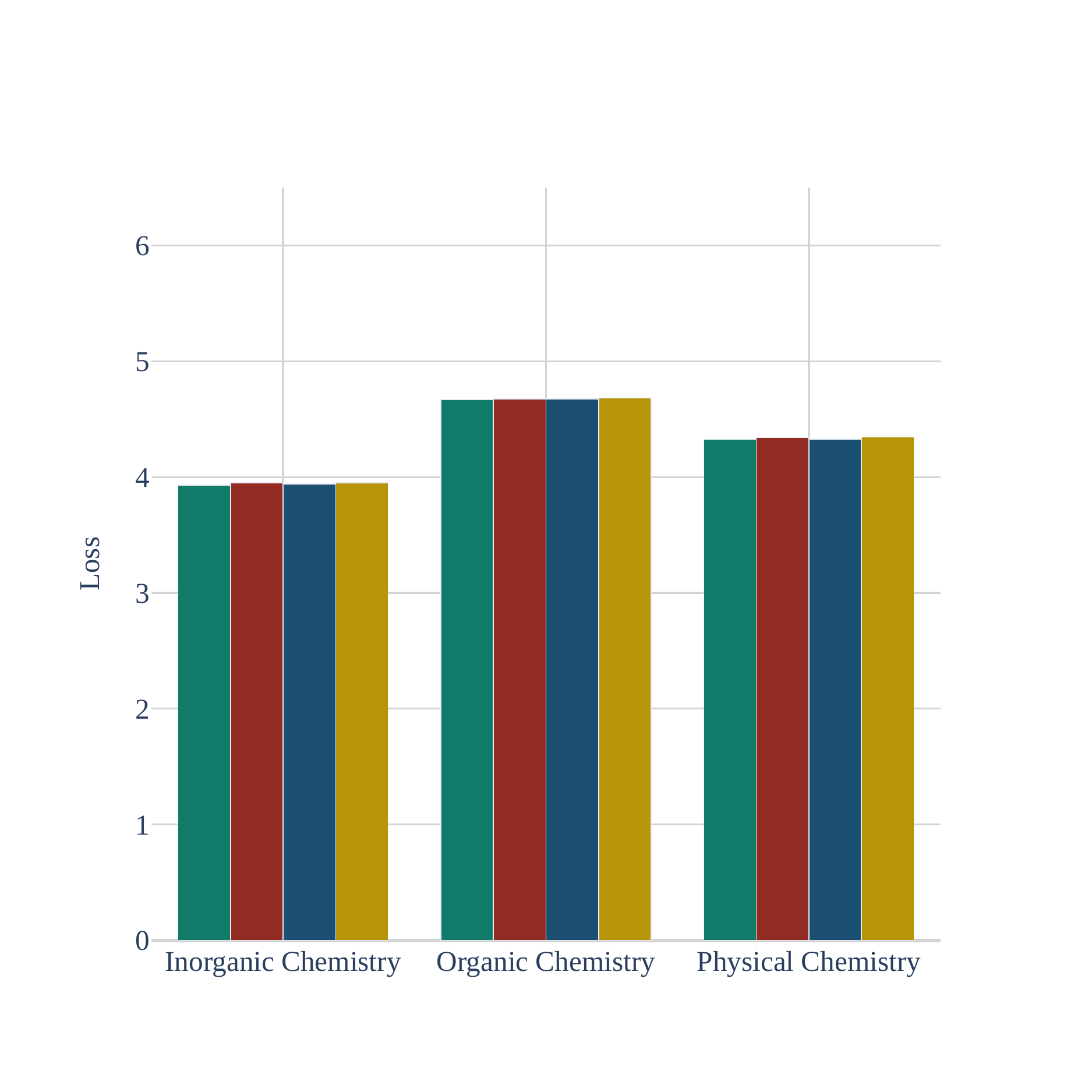}
  \label{fig:classwide-f2}}
   \caption{Intervention results on a Chemistry dataset with a clustered Gemma-1B model with $4$ clusters denoted by the four colors. Note that the non-modular model performs better than the modular model.}
  \label{fig:llm-chemistry}
\end{figure*}

Figure~\ref{fig:classwide-accuracies-both} compares class-wise accuracy of the model on the CIFAR-10 dataset with individual clusters turned OFF and ON respectively in a clustered model with an accuracy $>95\%$ for each label. 

In the top row we see there are a number of labels for which a cluster is able to perform well even when it is turned on in isolation. This means that the cluster is sufficient for performance on that label. 
Whenever performance plummets in the bottom row (when a specific cluster is turned off) this means that a specific cluster was necessary for performance.
For example, cluster 1 is both sufficient and necessary for classifying the label truck.




Future work may find that there are high-level semantic themes that different clusters learn. 
For example, the first cluster, which primarily learns predicting a ``ship'', also helps with ``airplane'' and ``bird'', all of which share a pointy front-end (nose or beak).

\subsubsection{Modules Do Not Specialize in Class-level Features for Both Modular and Non-Modular LLMs}
\label{sec:gemma-on-chem}

We perform two types of interventions on clusters to evaluate their contribution toward a model's computation:
\begin{enumerate}
    \item \textit{Type 1 / ON-intervention}: Here, we turn off (zero-ablate) the activations of all the other clusters, and run the model's forward pass using just the activations of a single cluster. This gives us a measure of the contribution of a cluster on its own toward predicting any given class or datapoint, which tells us whether a cluster is sufficient.
    \item \textit{Type 2 / OFF-intervention}: Here, we switch off a given cluster, and keep the activations of all the other clusters. This gives us a measure of how well all the other clusters combined can predict any given class or datapoint, which tells us whether a cluster is necessary.
\end{enumerate}

In Figure \ref{fig:llm-chemistry}, we show that optimizing the Gemma model for modularity on a Chemistry question-answering task with semantic categories \cite{XythicKChemistry} does not yield cluster-level specialization. We also find the performance to be worse for the modular model.

\subsubsection{GPT-2 Specializes on Specific Datapoints to Some Extent}
\label{sec:modular-llm}

\begin{figure}[t]
  \centering
  \subfloat
  {\includegraphics[width=\linewidth]{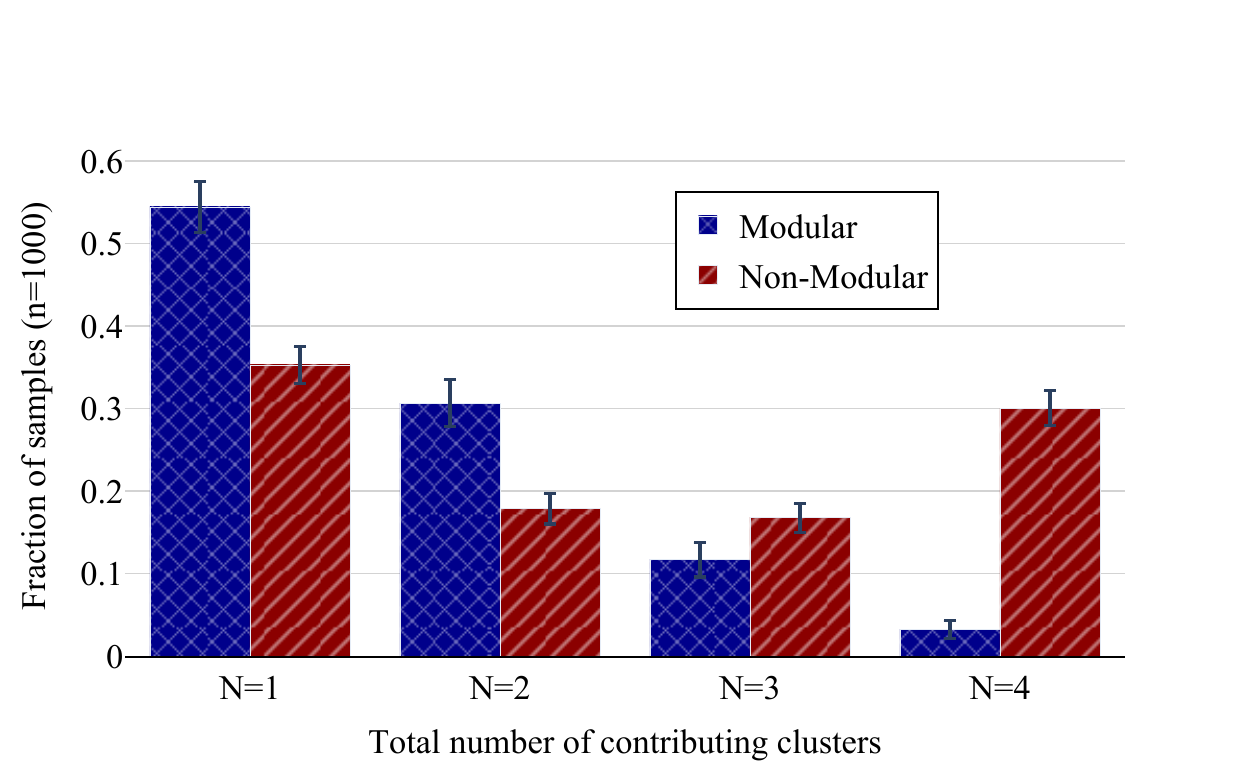}\label{fig:f1}}
  \caption{Fraction of samples for which there are exactly $k$ clusters that are individually able to correctly predict the answer. 
  Here $k\in[1,2,3,4]$ for a clustered model with $4$ clusters. 
  }
  \label{fig:llm-fractions}
\end{figure}

We also do an ON-intervention investigation on GPT2-small fine-tuned on wiki data.
In Figure \ref{fig:llm-fractions}, we see the fraction of samples for which there are exactly $k$ clusters that are individually able to correctly predict the answer. 
Note that in a modular model, most datapoints can only be solved by $1-2$ clusters, while $3-4$ modules can interchangably solve for a large fraction of datapoints in a non-modular model fine-tuned on the same data. 
See Appendix \ref{app:explanation-LLM-values} for a more elaborate explanation of these values.



\subsection{Individual Competency of Modules in LLMs}

\begin{figure*}[t]
  \centering
  \begin{subfigure}[t]{0.45\linewidth}
    \centering
    \includegraphics[width=\linewidth]{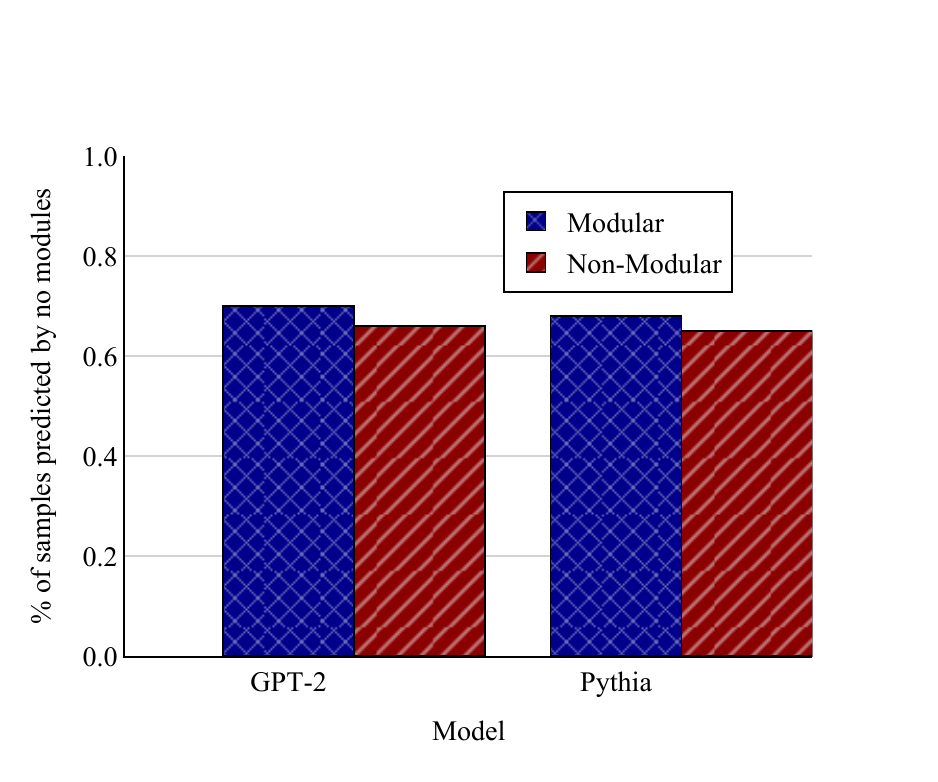}
    \caption{One module turned on in layer 5}
    \label{fig:llm-f2}
  \end{subfigure}
  \hfill
  \begin{subfigure}[t]{0.45\linewidth}
    \centering
    \includegraphics[width=\linewidth]{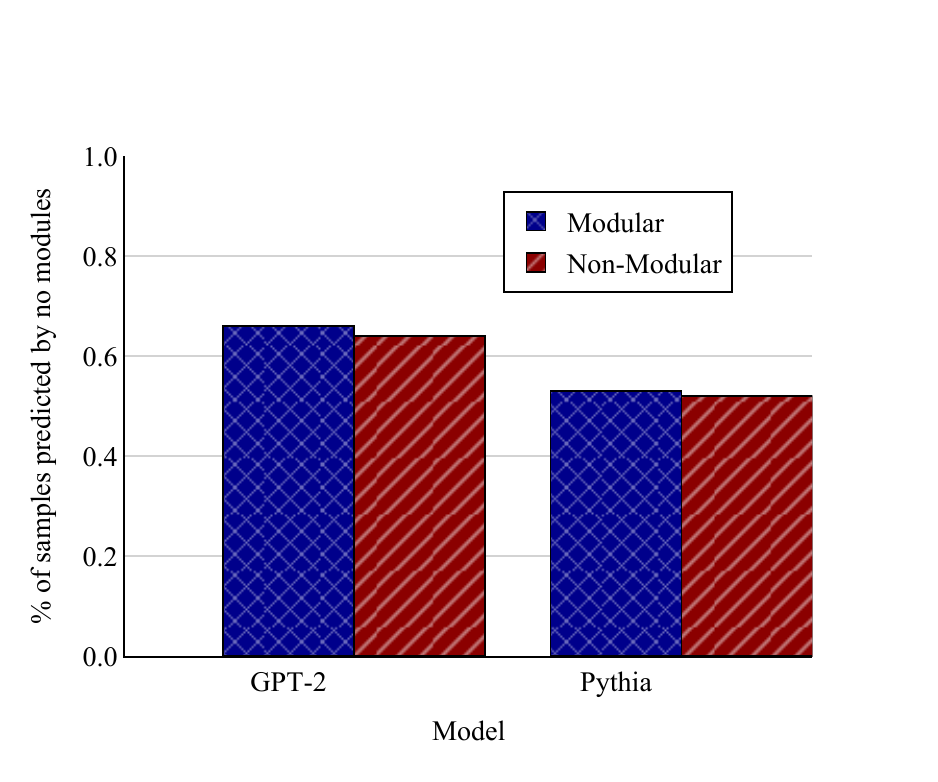}
    \caption{Three modules turned on in layer 5}
    \label{fig:llm-f1}
  \end{subfigure}
  \caption{Turning modules on or off in LLMs trained to be modular, or in LLMs where we arbitrarily select modules.}
  \label{fig:module-comparison}
\end{figure*}

We also check whether modules are able to independently solve datapoints in Figure \ref{fig:module-comparison}. 
Here we check for how many datapoints there were no modules that could solve the datapoint.
We find that for GPT-2 modular models had much more difficulty with predicting datapoints using a single module, as the number of incorrectly predicted samples was 80\%. 
For all other settings, modular and non-modular models behaved similarly, with modular modules being ever so slightly worse at predicting samples (using one or three modules).


\subsection{Modularity Encourages Smaller Circuits}\label{sec:circuit-size}

We use automated pruning to recursively remove edges that do not contribute (see Automated Circuit DisCovery \citep{conmy2023towards}) to extract the `effective circuit' for each label and define the ratio of the number of parameters in it to the number of parameters in the whole model to be the effective circuit size (ECS) for that label and model. A lower effective circuit size indicates that the model has learned a smaller task-specific circuit, and reduction in ECS is a proxy for interpretability gains. Since recursively removing edges is expensive, we only show results on simpler models, and leave more efficient pruning techniques for future work.

\begin{figure}[t]
    \centering
    \includegraphics[width=\linewidth,  trim=2 2 50 25, clip]{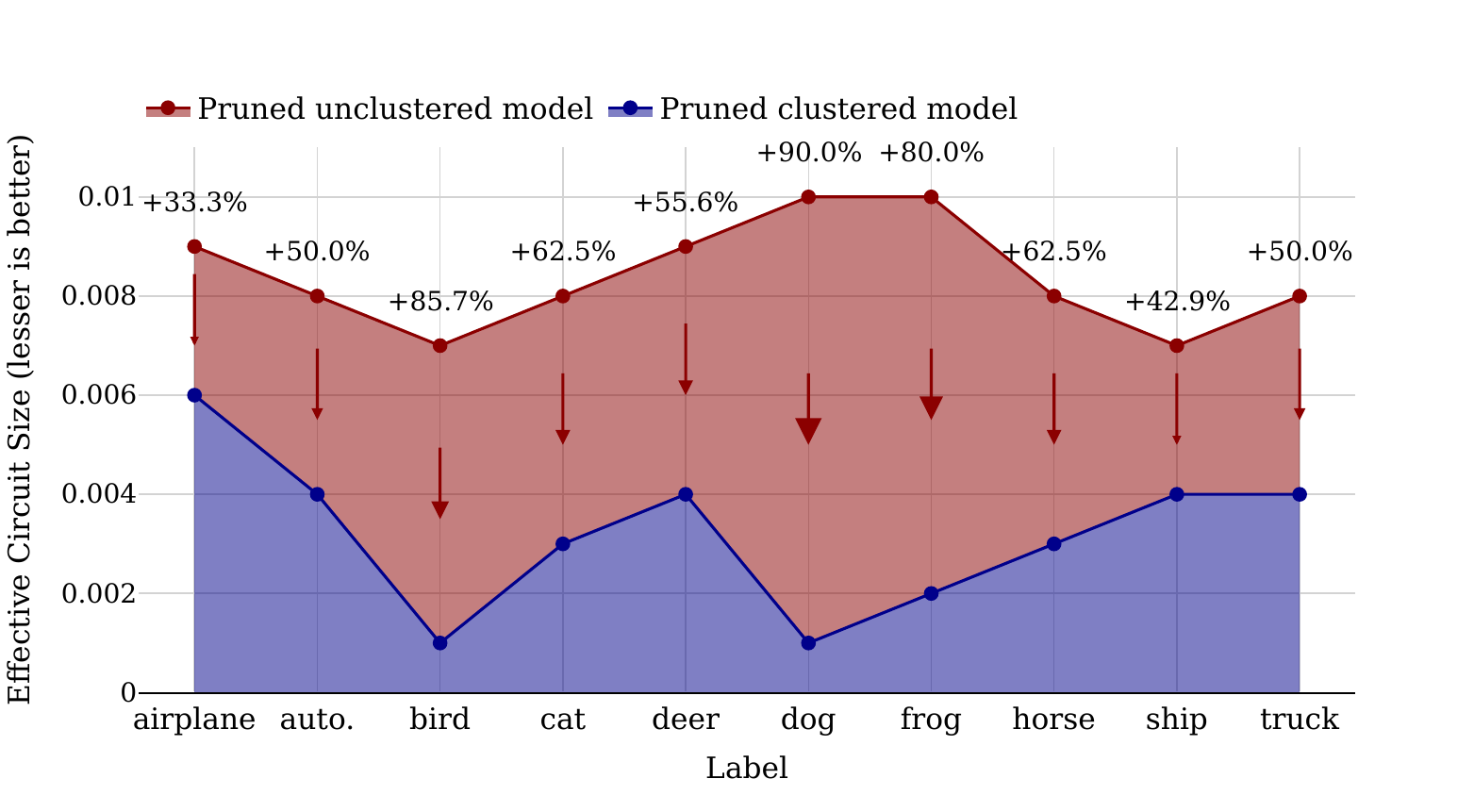}
    \caption{Percentage increase in Effective Circuit Size (ECS) for each label (as a fraction of the whole model) from the clustered to the unclustered models. Larger arrows denote a larger change in ECS.}
    \label{fig:ecs}
\end{figure}

Figure \ref{fig:ecs} compares the Effective Circuit Size (ECS) for each label for the clustered and unclustered models on CIFAR-10. 
We show that an unclustered model has, on average, $61.25\%$ more parameters in its effective circuits. 
In Appendix \ref{app:mnist}, we share similar results for the MNIST dataset \citep{deng2012mnist}.

Our clusterability measure regularizes against connections between modules. 
In the 271424 parameter CNN model we use, there are 1024 such connections.
So even if we all of these connections were zero, that would only be a reduction of 0.37\% of the total number of parameters. 
This means that the reduction in parameters in effective circuits is unlikely due to regularization leading to a reduction in non-zero parameters, and instead due to underlying restructuring of the circuits.

\section{Theoretical Analysis of the Effects of Modularity}\label{sec:theoretical}

\subsection{Polytopes in a Modular Network}

We show that the partitioning of neural networks into polytopes on which the network is linear \citep{sudjianto2020unwrapping} becomes coarser when we make a model modular. In other words, the simplicity of the function space of a network increases by adding modularity constraints. In this section, we restrict to fully connected ReLU feed-forward networks. 

\begin{definition}[ReLU Network]
A ReLU network $\mathcal{N} \colon \mathbb{R}^n \rightarrow \mathbb{R}^m$, is a composition of $L \in \mathbb{N}$ hidden layers given by $\chi^{(l)} = \sigma( {W^{(l)}} \chi^{(l-1)} + b^{(l)})$, where $\sigma$ is an element-wise ReLU activation function, $\sigma(x_i) = \max\{0,x_i\}$. We define $\chi^{(0)} = x$ and the output layer is given by 
\[\mathcal{N}(x) = W^{(L+1)} \chi^{(L)} + b^{(L+1)}.\] 
The number of neurons in each layer is given by a vector $\mathbf{N} = [n_1, n_2, \ldots, n_L]$, and all activations are in the positive reals, i.e. $\chi^{(l)} \in \mathbb{R}^{n_l}_{\geq 0}$ for all $l \in \{1,\ldots,L\} = [L]$.
\end{definition}

ReLU networks can be described as a collection of linear models that are applied to certain regions of the input space \citep{sudjianto2020unwrapping}. These regions partition the input space $\mathbb{R}^n$. 

\begin{proposition}\label{prop:network-to-partition}[\citet{sudjianto2020unwrapping}]
For a ReLU network $\mathcal{N} \colon \mathbb{R}^n \rightarrow \mathbb{R}^m$ there is a finite partition $\Omega$ of $\mathbb{R}^n$ of cardinality $p := \# \Omega$ such that for each part $\omega \in \Omega$ there exists a piece-wise linear function $f\colon \mathbb{R}^n \rightarrow \mathbb{R}^m$, and its restriction on $\omega$, denoted $f|_\omega$, can be described by a linear function: 
\[f|_\omega (x) = \alpha_\omega^T x + \beta_\omega.\]
Moreover, each part is a  polytope, given by the intersection of a collection of half-spaces.
\end{proposition}

In some sense, the number of polytopes gives a measure of the granularity or refinement of a network.
Given a network with neurons $\mathbf{N} = [n_1, n_2, \ldots, n_L]$, 
there are at most $2^{n_1} \times 2^{n_2} \times \ldots \times 2^{n_L}$ polytopes. Now suppose that the weight matrix $W^{(l)}$ in layer $\chi^{(l)}$ is modular and contains $k$ modules, meaning that we can divide $n_{l-1}$ and $n_{l}$ into $k$ sets $[n_{l-1}^1, \ldots, n_{l-1}^k]$ and $[n_{l}^1, \ldots, n_{l}^k]$ such that the only non-zero weights in $W^{(l)}$ are between $n_{l-1}^i$ and $n_{l}^i$.
In this case the number of half-space conditions generated by $\chi^{(l-1)}$ and $\chi^{(l)}$ is no longer $2^{n_{l-1}} \times 2^{n_l}$, but it now is $2^{n_{l-1}} \times 2^{n_{l}^1} + \ldots + 2^{n_{l-1}} \times 2^{n_{l}^k}$, which is substantially fewer.
This means that modularity increases the simplicity of the function-space of a network, which comes at the cost of expressivity.

\subsection{Concept Representations in Activation Space}\label{sec:theoretical-representations}

Based on the Johnson-Lindenstrauss lemma we will show that the number of nearly orthogonal points that can be represented in a layer goes down when we make the layer modular.

\begin{proposition}[\citet{johnson1984extensions}]
Let $0< \epsilon < 1$ and let $X$ be a set of $m$ points in $\mathbb{R}^N$ and $n > \frac{8(ln(m))}{\epsilon^2}$,
then there exists a linear map $f \colon \mathbb{R}^N \rightarrow \mathbb{R}^n$ such that for all $u,v \in X$ we have
\[ (1- \epsilon) \| u - v\|  \leq \| f(u) - f(v) \|^2 \leq (1+ \epsilon) \| u - v \|^2. \]
\end{proposition}

In other words, a set of $m$ nearly orthogonal points in $\mathbb{R}^N$ can be projected onto nearly orthogonal directions in a much smaller dimensional space $\mathbb{R}^n$,
as long as $n > \frac{8(ln(m))}{\epsilon^2}$ or equivalently $ e^{\frac{ \epsilon^2 }{8} \cdot n} > m $.
This means that the number of points that can be projected orthogonally onto $\mathbb{R}^n$ is exponential in $n$.

Given a network with neurons $\mathbf{N} = [n_1, n_2, \ldots, n_L]$ where the weight matrix $W^{(l)}$ in layer $\chi^{(l)}$ is modular and contains $k$ modules such $n_{l-1}$ and $n_{l}$ can be divided into $k$ sets $[n_{l-1}^1, \ldots, n_{l-1}^k]$ and $[n_{l}^1, \ldots, n_{l}^k]$ such that the only non-zero weights in $W^{(l)}$ are between $n_{l-1}^i$ and $n_{l}^i$.

Then the number of nearly orthogonal points that can be represented in layer $n_l$ becomes $~ e^{n_{l}^1} + \ldots + e^{n_{l}^k} $ as opposed to $~ e^{   {n_{l}^1} + \ldots + {n_{l}^k} }$. 
This means that as a result of modularizing the weight matrix $W^{(l)}$, the model learns to perform computation by using features from an exponentially smaller set of subspaces. Sparse Autoencoders (SAEs) \cite{cunningham2023sparseautoencodershighlyinterpretable} extract linear subspaces as latents for human-interpretable features, and modularity helps reduce the search space for SAEs exponentially. Training SAEs on modular models and modules is an interesting direction for future work.

\section{Conclusion}

We show that a simple regularizer can split a neural network's layer into non-interacting modules, but these modules do not yield the interpretability benefits that were expected. 
We show that the average circuit size and the search space improve with more modules without a decrease in overall performance (for the models and tasks that we investigated).


Future work could investigate if there are alternative regularizers that do lead to clear interpretability benefits, and task specialization on harmful or desired tasks.

One possible hypothesis why our modularity metric does not work is that it operates on the wrong level of granularity, and further exploration into this would be an important future direction.
To resolve this, future work could build an automated subtask detection mechanism, whereby the subtasks do not have to be specified beforehand.

\bibliography{main}

\appendix

\section{Algorithm}\label{app:algorithm}

\begin{algorithm}[t]
   \caption{Bipartite Spectral Graph Clustering (BSGC)}
   \label{alg:spectral_clustering}
   \begin{algorithmic}[1]
      \STATE \textbf{Input:} Similarity matrix $A$ ($m \times n$), number of clusters $k$
      \STATE \textbf{Output:} Bipartite clusters $\text{U, V}$ of input/output neurons
      \STATE \textbf{1.} Compute normalized similarity matrix:
      \STATE \quad $D_U, D_V \gets \text{diag}\left(\sum_i A_{i, \cdot}\right), \text{diag}\left(\sum_j A_{\cdot, j}\right)$
      \STATE \quad $\tilde{A} \gets D_U^{-1/2} A D_V^{-1/2}$
      \STATE \textbf{2.} Perform Singular Value Decomposition (SVD):
      \STATE \quad $U, \Sigma, V^T \gets \text{SVD}(\tilde{A}, k)$
      \STATE \textbf{3.} Perform KMeans clustering:
      \STATE \quad $\text{U}, \text{V} \gets \text{KMeans}(k, U), \text{KMeans}(k, V^T)$
      \STATE \textbf{Return:} Clusters $\text{U}, \text{V}$
   \end{algorithmic}
\end{algorithm}

\section{Hyperparameters}\label{app:hyper}

We list the hyperparameters and various design choices in our experiments in Tab. \ref{tab:hyperparameters}, and refer to our codebase for more details (to be added during de-anonymization).

\begin{table}[ht]
\centering
\caption{Hyperparameter choices and other experiment details.}
\vspace{5pt}
\begin{tabular}{l l}
\toprule
\textbf{Hyperparameter/Design Choice} & \textbf{Value} \\
\midrule
Dataset & CIFAR-10 (or MNIST) \\
Batch Size & 64 \\
\midrule
Optimizer & Adam \\
Learning Rate ($\alpha$) & $1 \times 10^{-3}$ \\
Criterion & Cross-Entropy \\
Clusterability Factor & 20 \\
Number of Clusters & 4 \\
\midrule
\textbf{MLP (for MNIST)} & \\
\midrule
Input Size & $28 \times 28$ \\
Hidden Layer Sizes & 64, 64 \\
Activation Function & ReLU \\
Output Size & 10 \\
Bias & False \\
\midrule
\textbf{CNN (for CIFAR-10)} & \\
\midrule
Input Channels & 3 \\
Conv Layer 1: Out Channels & 16 \\
Conv Layer 1: Kernel Size & 3 \\
Conv Layer 1: Stride & 1 \\
Conv Layer 1: Padding & 1 \\
FC Layer 1: Input Features & $16 \times 16 \times 16$ \\
FC Layer 1: Output Features & 64 \\
FC Layer 2: Output Features & 64 \\
Output Size & 10 \\
Bias & False \\
Activation Function & ReLU \\
\midrule
\textbf{Transformer (for Modular Arithmetic)} & \\
\midrule
Learning Rate ($\alpha$) & $1 \times 10^{-3}$ \\
Weight Decay & 1.0 \\
Prime Number ($p$) & 113 \\
Model Dimension ($d_{\text{model}}$) & 128 \\
Function Name & Addition \\
Training Fraction & 0.3 \\
Number of Epochs & 200 \\
Number of Layers & 1 \\
Batch Style & Full \\
Vocabulary Size ($d_{\text{vocab}}$) & $p + 1$ \\
Context Length ($n_{\text{ctx}}$) & 3 \\
MLP Dimension ($d_{\text{mlp}}$) & $d_{\text{model}}$ \\
Number of Heads & 4 \\
Activation Function & ReLU \\
Device & CUDA \\
Use LayerNorm & False \\
\midrule
Pruning Method & Iterative weight pruning \\
Pruning Criteria & Performance-based (loss and accuracy) \\
\midrule
Effective Circuit Size Calculation & Fraction of non-zero weights \\
\bottomrule
\end{tabular}
\label{tab:hyperparameters}
\end{table}

\section{Results on MNIST}\label{app:mnist}

\begin{figure}[h]
\centering
    \includegraphics[width=0.4\textwidth,  trim=2 2 40 80, clip]{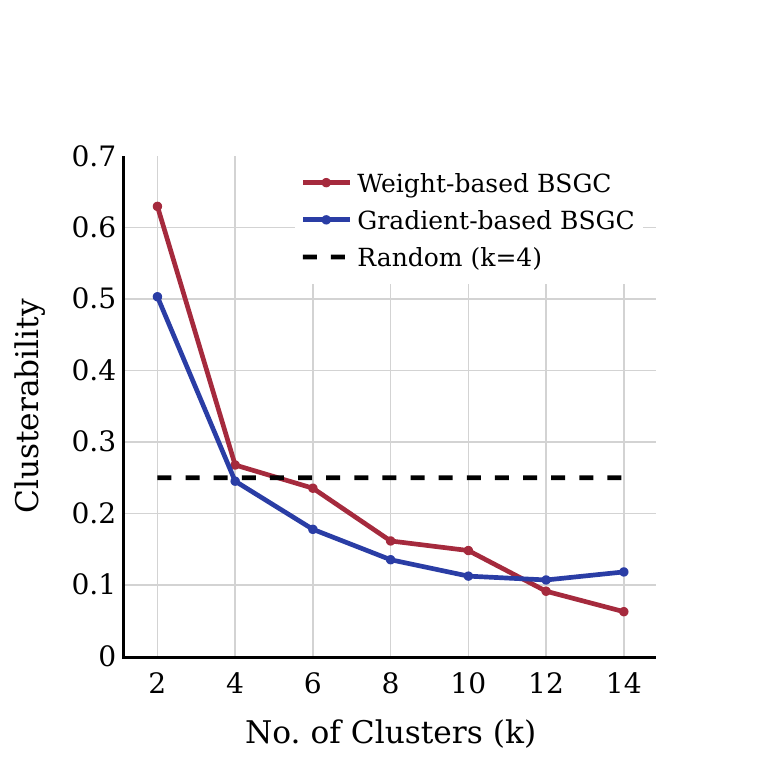}
    \caption{Clusterability (enmeshment) of the model with $k$ bipartite clusters using Algorithm~\ref{alg:spectral_clustering} on MNIST.}
    \label{fig:mnist_clusterability}
\end{figure}

\begin{figure}[h]
    \centering
    \includegraphics[width=0.8\linewidth,  trim=2 2 300 100, clip]{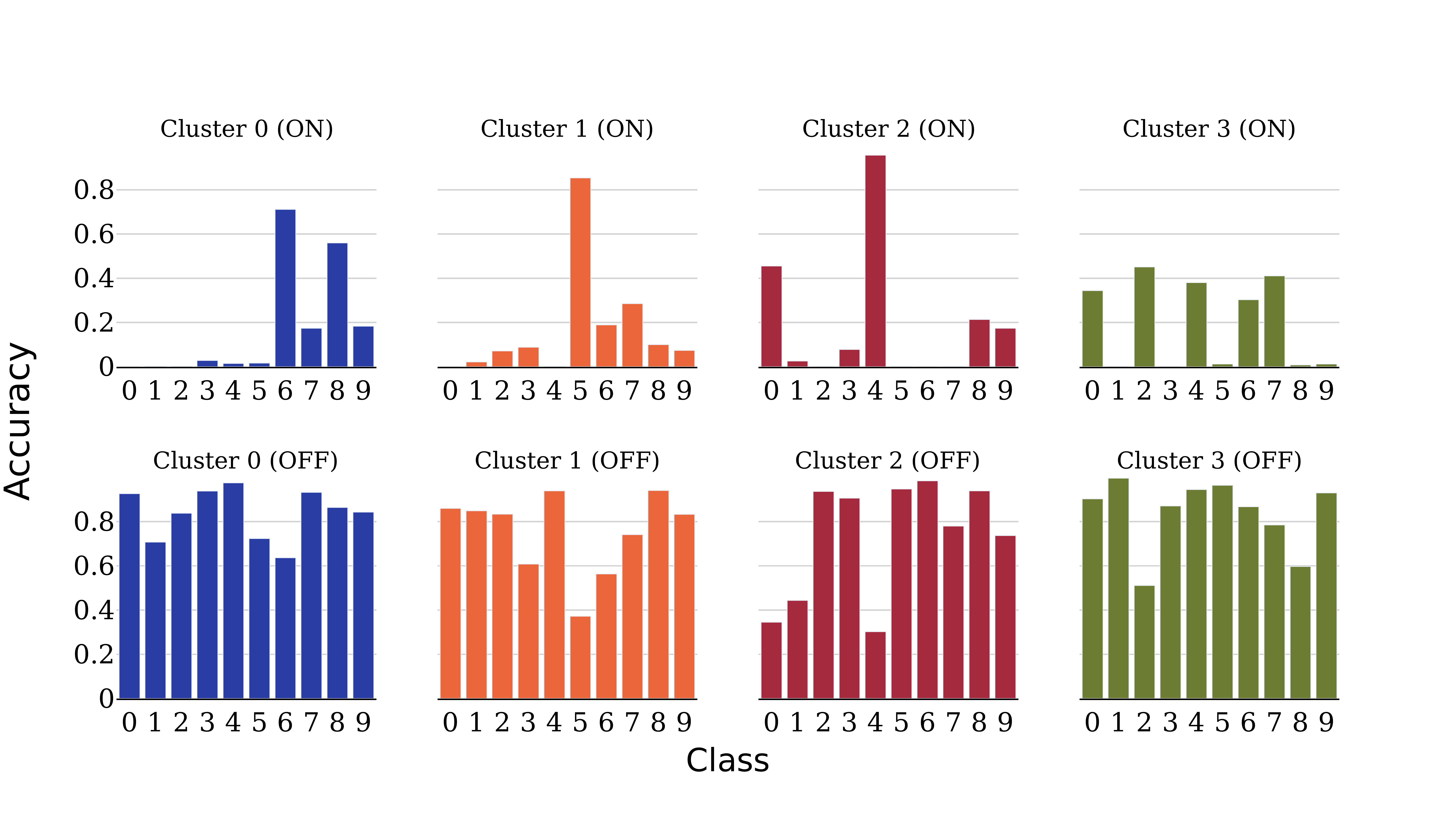}
    \caption{Class-wise accuracy for each label with clusters turned ON (top) and OFF (bottom) for MNIST. Note individual clusters learning near-complete circuits for various labels.}
    \label{fig:mnist_classwise-accuracies}
\end{figure}

Here we present our results on the MNIST \citep{deng2012mnist} dataset. For the clusterability of the model with $k$ bipartite clusters, see Fig. \ref{fig:mnist_clusterability}, and for the class-wise accuracies for each label with clusters turned ON and OFF, see Fig. \ref{fig:mnist_classwise-accuracies}.

\begin{figure}[h]
    \centering
    \includegraphics[width=0.8\linewidth,  trim=2 2 50 25, clip]{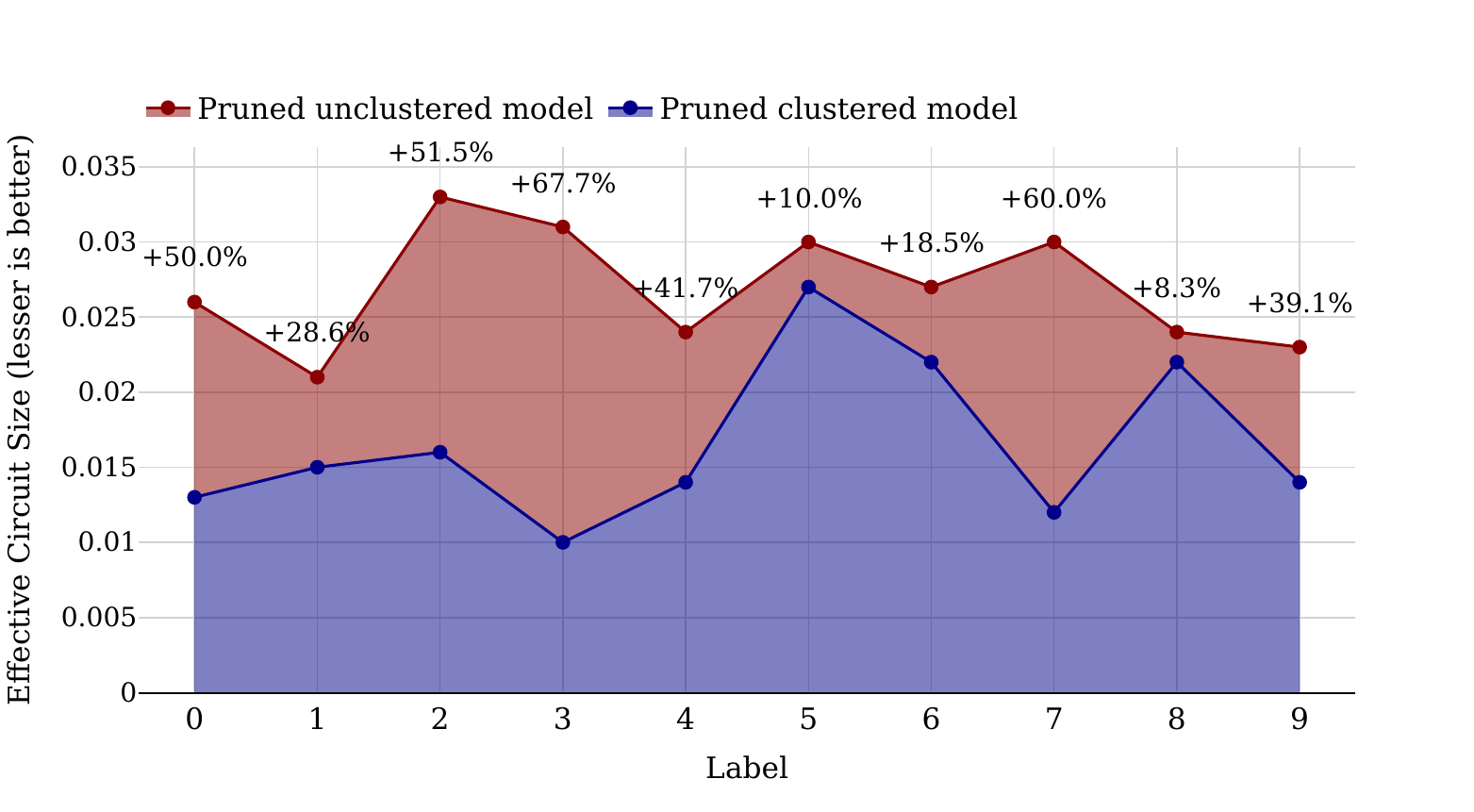}
    \caption{Effective Circuit Size (ECS) of circuits for each label as a fraction of the whole model for both clustered and unclustered models trained on MNIST. Larger arrows denote a larger reduction in ECS.}
    \label{fig:mnist_ecs}
\end{figure}

\begin{figure}[t]
\centering
\includegraphics[width=0.9\columnwidth, trim=10 10 40 50, clip]{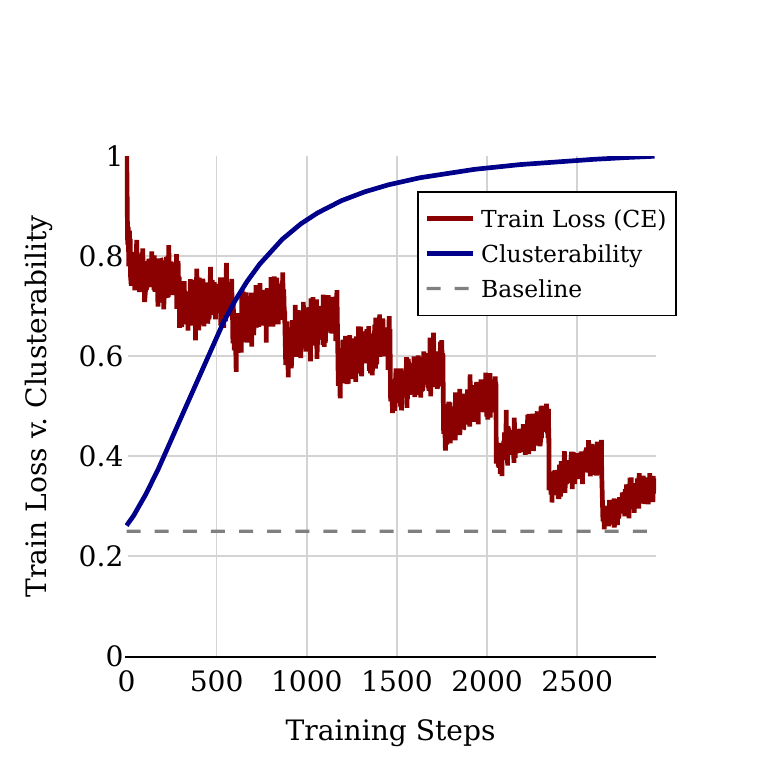}
\caption{Combined clusterability of the MLP input of all layers in GPT2-small and the cross-entropy training loss on a subset of the Wikipedia dataset during modularity training.}
\label{fig:wiki-training}
\end{figure}

\begin{figure*}[h]
\centering
    \includegraphics[width=0.8\textwidth,  trim=2 2 40 80, clip]{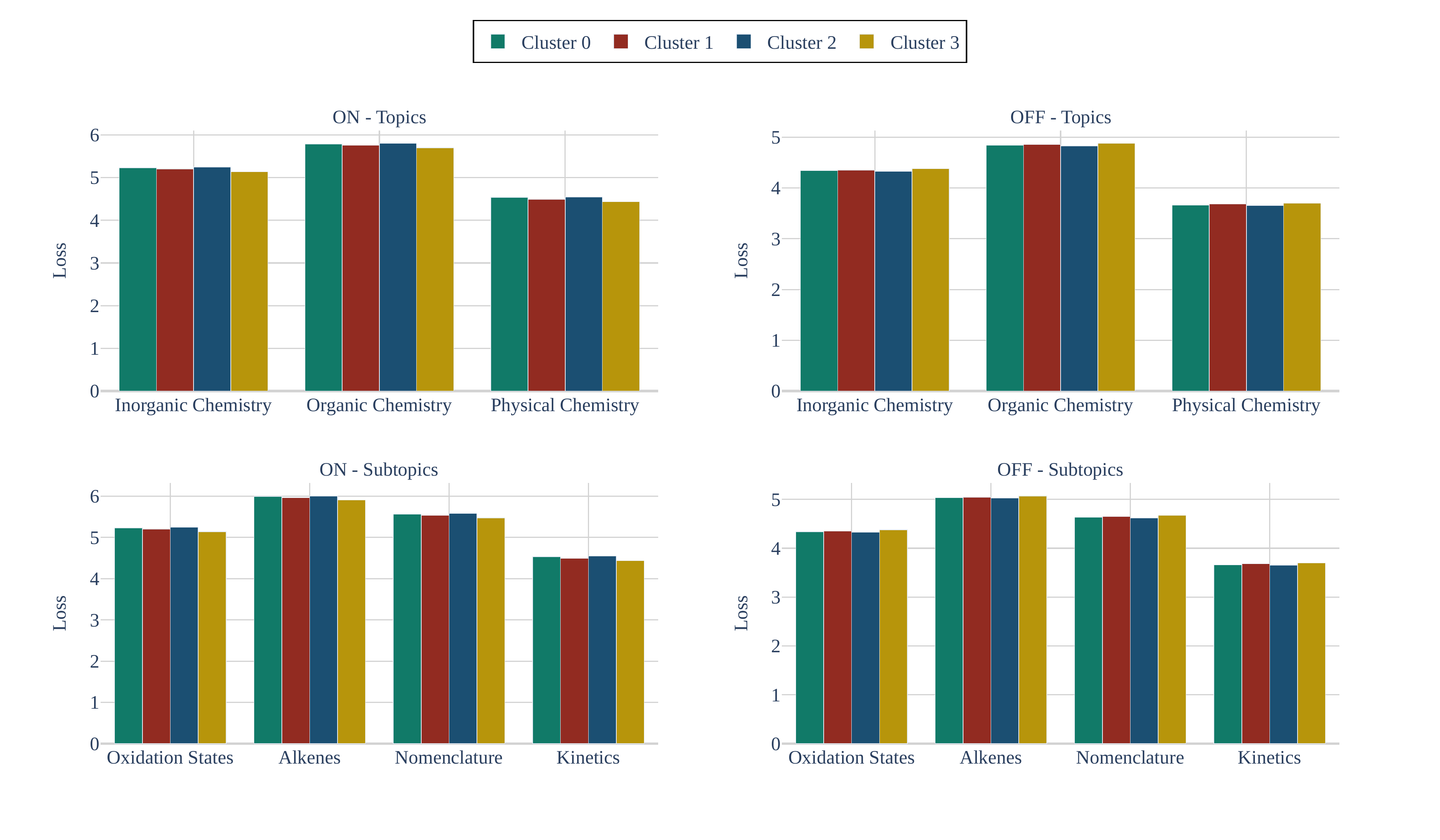}
    \caption{Intervention results on LLMs on Chemistry (Modular) for two levels of semantic clustering.}
    \label{fig:chemistry-app-modular}
\end{figure*}

\begin{figure*}[h]
\centering
    \includegraphics[width=0.8\textwidth,  trim=2 2 40 80, clip]{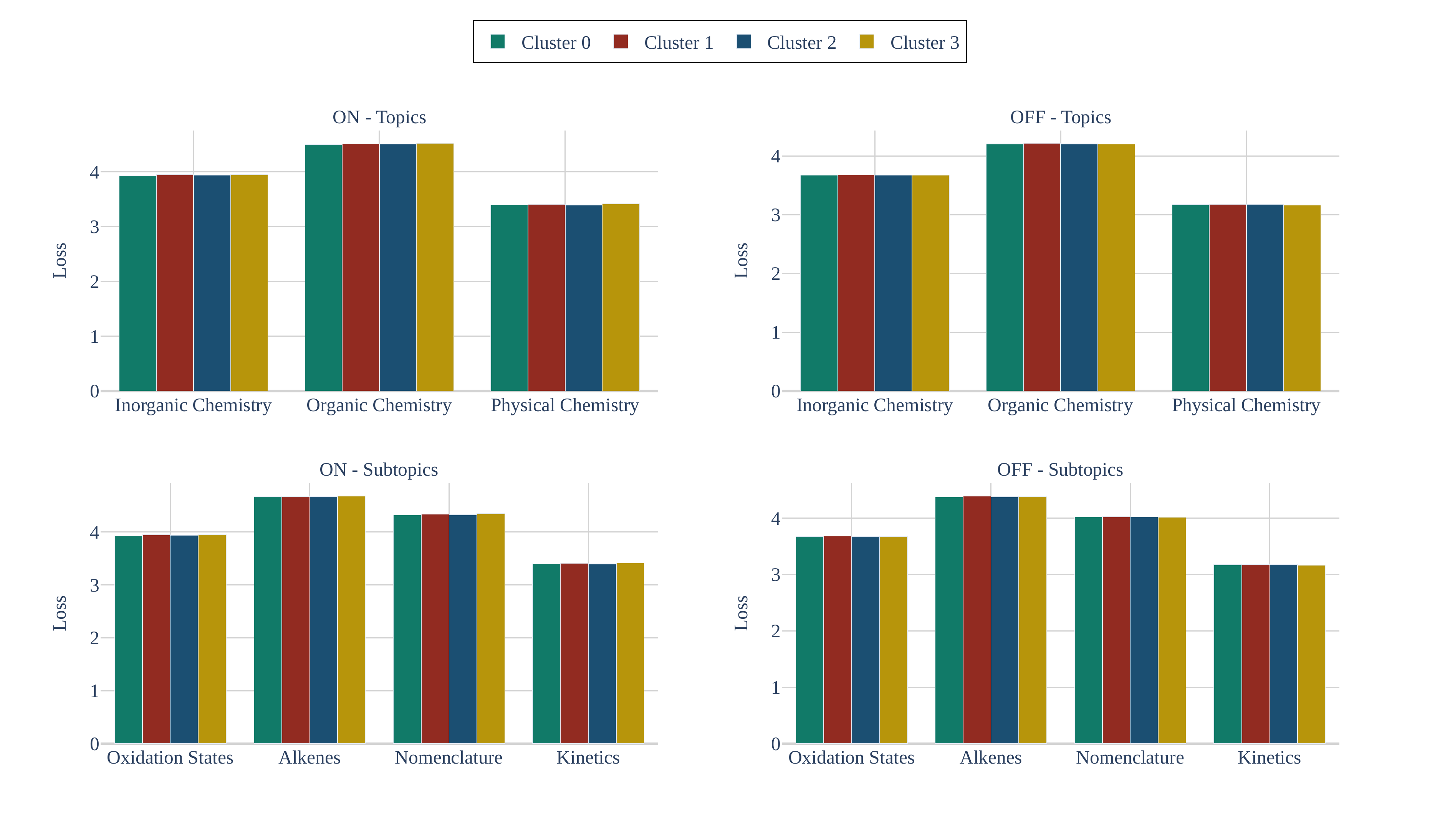}
    \caption{Intervention results on LLMs on Chemistry (Non-modular) for two levels of semantic clustering.}
    \label{fig:chemistry-app-non-modular}
\end{figure*}

\section{Clustered Layer Visualization}

\begin{figure}[h]
    \centering
    \includegraphics[width=0.8\linewidth]{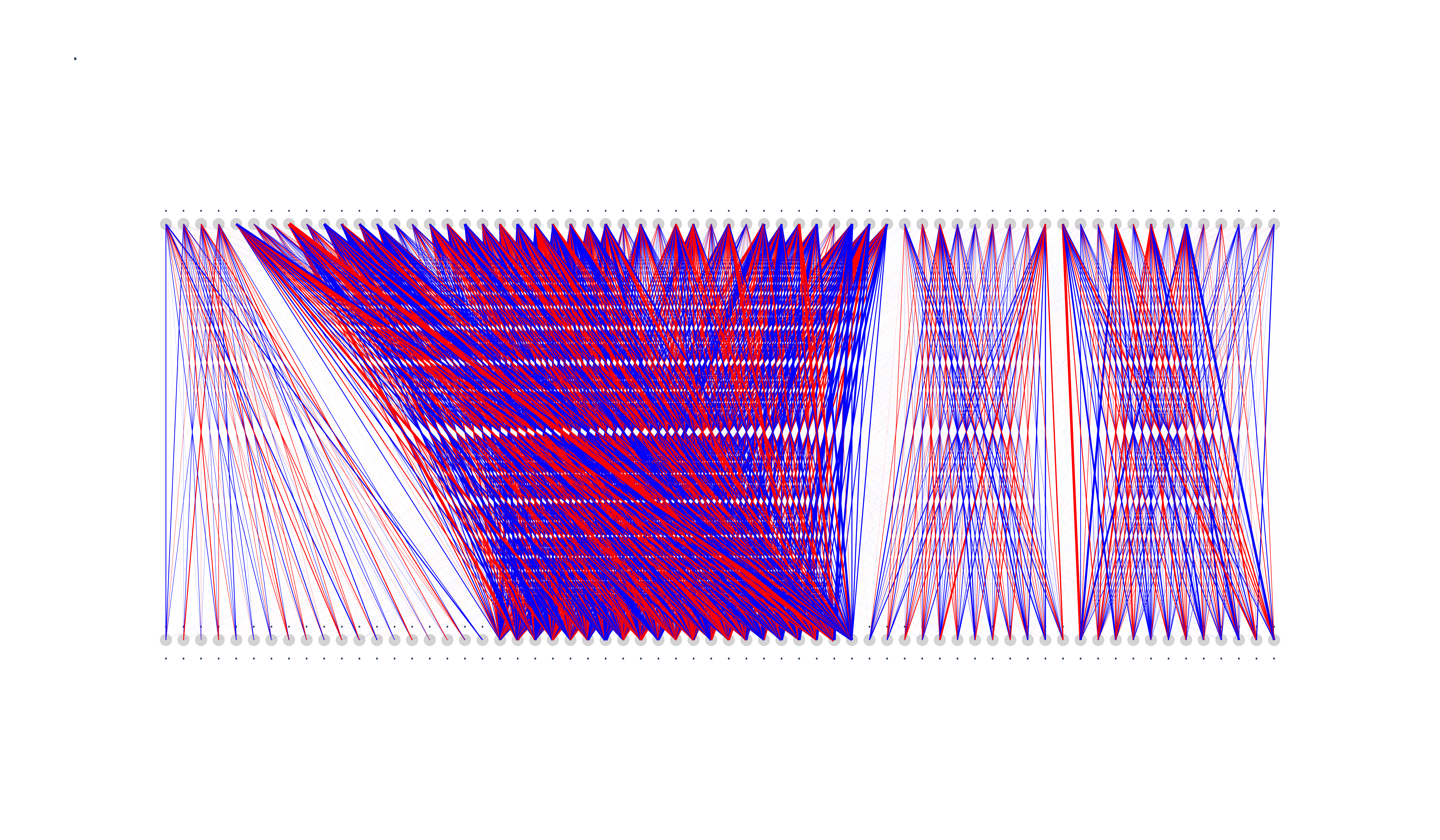}
    \caption{Visualizing the clustered layer $fc2$ of the model. Red and blue denote negative and positive weights respectively.}
    \label{fig:cluster-visualization}
\end{figure}

Fig. \ref{fig:cluster-visualization} shows the visualization of the fully connected layer trained on CIFAR10 of the network with bipartite clusters.

\section{Interpretability of Modular Transformer for Modular Arithmetic}\label{app:arithmetic}

\begin{figure}[h]
    \centering
    \includegraphics[width=\linewidth]{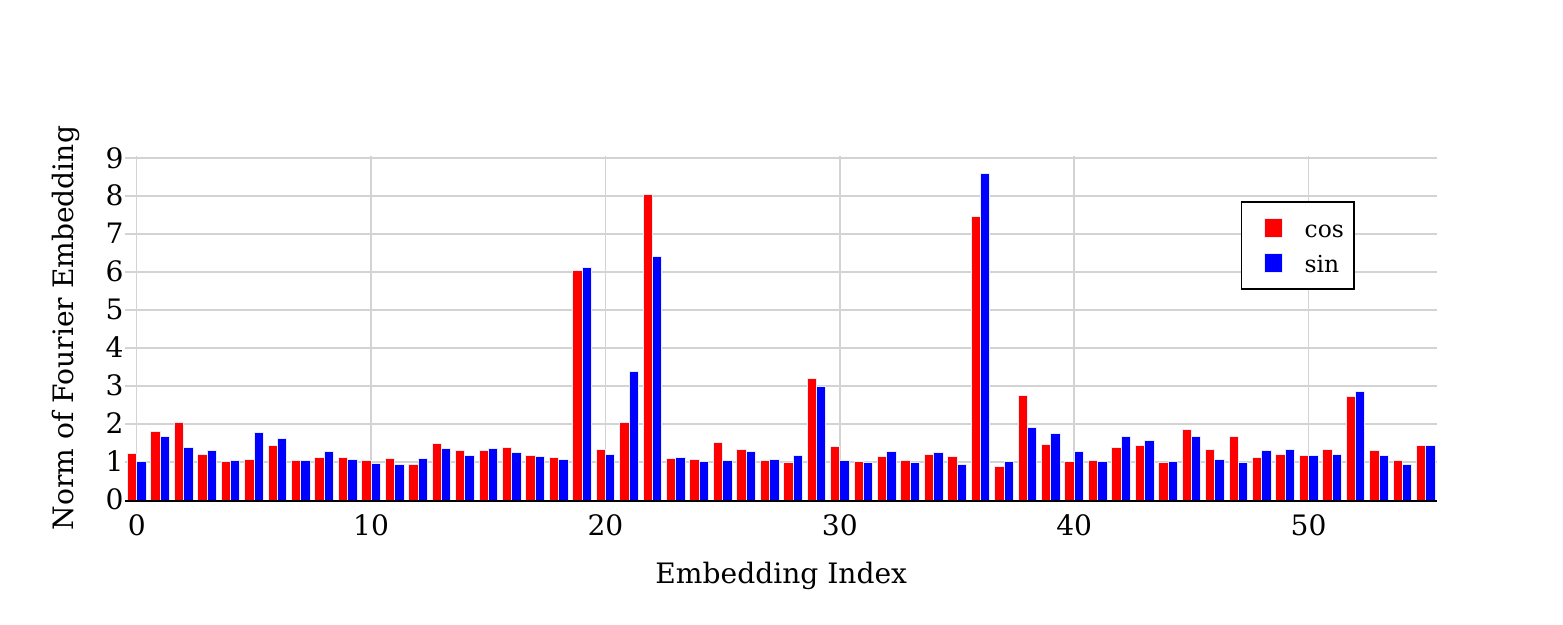}
    \caption{Norms of the Fourier embeddings for sine and cosine functions.}
    \label{fig:fourier-norms}
\end{figure}

In Fig. \ref{fig:fourier-norms}, we show the $L2$ norms of the embeddings with the discrete Fourier transformation are very similar to the observations made in \citet{nanda2023progressmeasuresgrokkingmechanistic}, indicating that the circuits in the modular model learn the same algorithm to perform modular addition. 

However, interpreting such a model is much easier with modularity, because modular MLPs have a much simpler representation space, and modular Attention matrices imply a $n$ times simpler circuit search space as compared to a non-modular model in the presence of $n$ completely non-intersecting clusters with a clusterability value of $1$, as we observe in \ref{sec:modular-modular-addition}.

\section{Type-1 and Type-2 Intervention Performance}

\begin{figure}[h!]
    \centering
    \includegraphics[width=\linewidth]{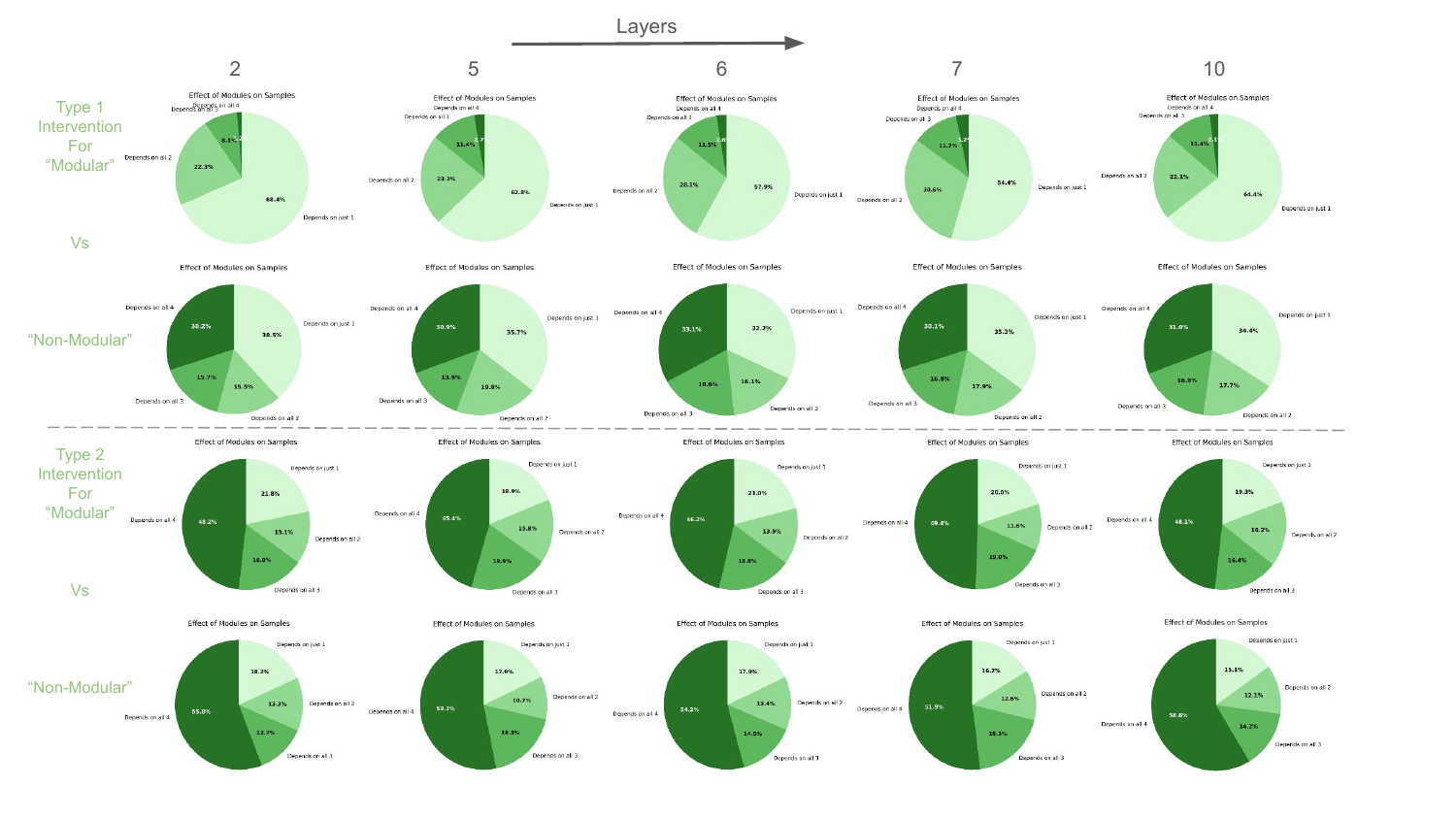}
    \caption{Type $1$ and Type $2$ interventions illustrating the dependence of samples on individual or pairs of modules. The figure highlights a significant difference in sample dependence between modular and non-modular models, particularly in Type $1$ interventions. Approximately $30\%$ of samples in the non-modular model rely on all $4$ modules for correct predictions, compared to only about $1\%$ in the modular model. Similarly, in Type $2$ interventions, there is a notable $10\%$ difference in module dependence on all $4$ samples between the two models.}
    \label{fig:modvsnmod}
\end{figure}

Figure \ref{fig:modvsnmod} shows Type $1$ and Type $2$ interventions illustrating the dependence of samples on individual or pairs of modules. The figure highlights a significant difference in sample dependence between modular and non-modular models, particularly in Type $1$ interventions. Approximately $30\%$ of samples in the non-modular model rely on all $4$ modules for correct predictions, compared to only about $1\%$ in the modular model.

\section{Explanation of LLM Intervention Values}\label{app:explanation-LLM-values}

To calculate these values, we 
(1) only consider the subset of the test set for which the model without intervention gets hundred percent accuracy; and (2) for each module we do a type 1 intervention (i.e. only keeping that module on and turning the other three off) and keep track of the datapoints that are now handled incorrectly.
If a datapoint is handled incorrectly when only module X is turned on, 
then module X is not sufficient on its own. 
We then count for how many datapoints there was one module (N=1) such that this module is not sufficient on its own, for how many datapoints there were two modules (N=2) such that this module is not sufficient on its own, and so on.

In Figure \ref{fig:llm-fractions} we find that for modular models N=1 is higher, which means there were many datapoints for which there was only one module that was not sufficient on its own, i.e. many modules (three) were able to handle the datapoint correctly when they were turned on in isolation.
For non-modular models N=4 is higher, which means that there were more datapoints such that all modules were not sufficient on their own, i.e. these datapoints required two or more modules to be turned on. 


\section{Clusterability for Pythia Models for Different Number of Clusters}
\label{app:pythia}

In Fig. \ref{tab:clusterability-diff-n}, we share the clusterability for pythia-70m on different values of $k$ (number of clusters).

\begin{table}[h]
    \centering
    \begin{tabular}{c|c|c|c}
        \hline
        \textbf{k} & \textbf{Baseline} (\( 1/k \)) & \textbf{Before Finetuning} & \textbf{After 2 Finetuning Epochs} \\
        \hline
        2 & 0.5000 & 0.6009 & 0.9833 \\
        4 & 0.2500 & 0.2403 & 0.9658 \\
        5 & 0.2000 & 0.2109 & 0.9629 \\
        6 & 0.1667 & 0.1611 & 0.9548 \\
        8 & 0.1250 & 0.1201 & 0.9463 \\
        \hline
    \end{tabular}
    \caption{Clusterability scores before and after finetuning for different numbers of clusters for Pythia-70m. Baseline is given by \( 1/k \).}
    \label{tab:clusterability-diff-n}
\end{table}

\section{Chemistry on Gemma}

\begin{figure*}[h]
  \centering
  \includegraphics[width=0.9\linewidth, trim={2 2 50 25}, clip]{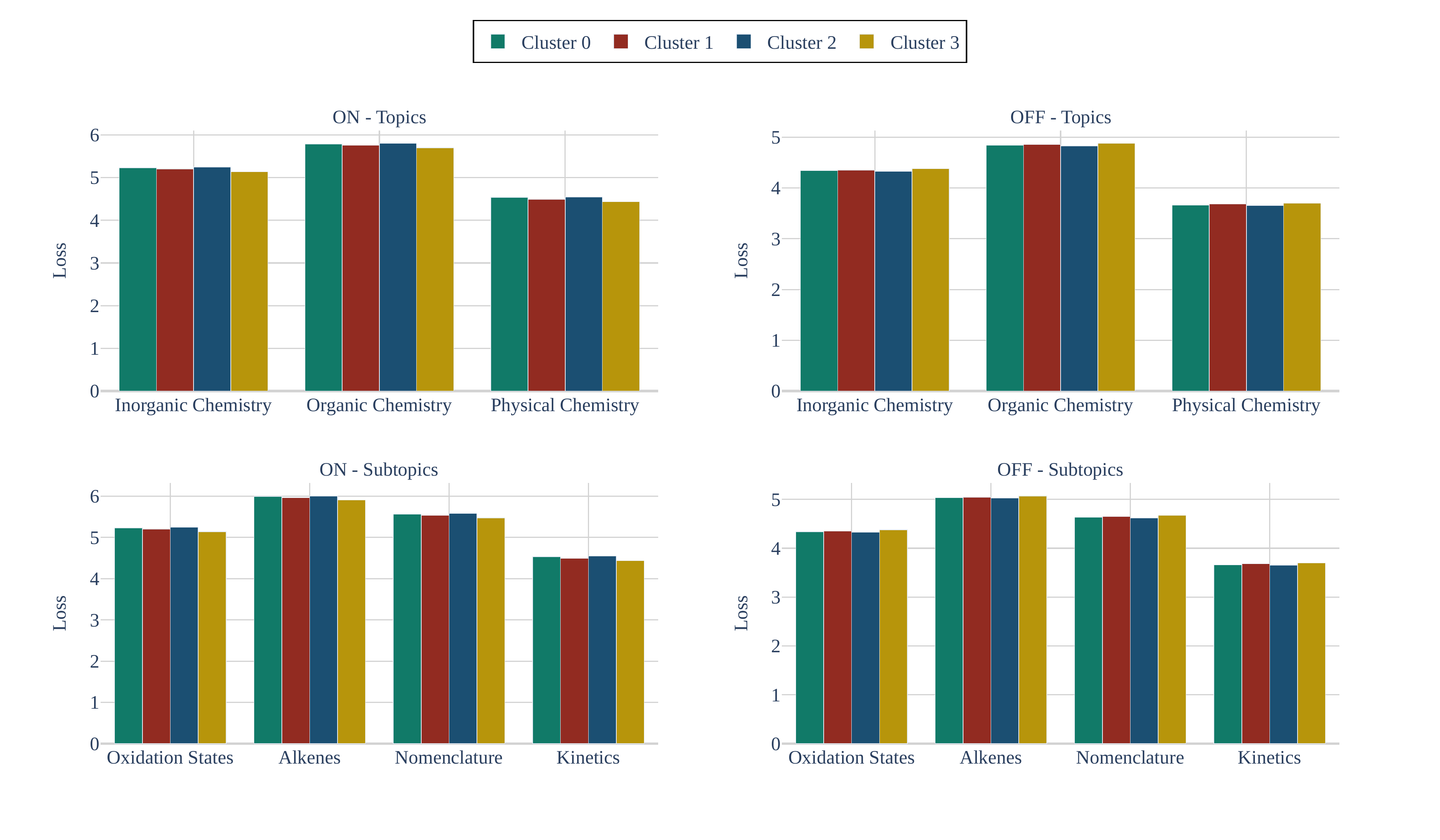}
  \caption{Intervention results on a Chemistry dataset with a clustered Gemma-1B model with $4$ clusters.}
  \label{fig:llm-chemistry}
\end{figure*}

\section{Details on Other Metrics}

\subsection{Community Structure}\label{sec:community-structure}

In an undirected and unweighted graph (where a weight can only take on the values 0 and 1), the expected number of edges between two nodes $i$ and $j$ is
\begin{align*}
& E[J_{ij}] = \frac{ \sum\limits_{i=1}^{n} W_{ij} \sum\limits_{j=1}^{n} W_{ij}}{2 \sum\limits_{i=1}^{n} \sum\limits_{j=1}^{n} W_{ij}}, \text{ the community structure is } \\
& Q = \frac{ \sum\limits_{i=1}^{n} \sum\limits_{j=1}^{n} \big( W_{ij} -  E[J_{ij}] \big) \cdot \mathbb{I}_{(i \in C_U(u) \land j \in C_V(u))} }{2 \sum\limits_{i=1}^{n} \sum\limits_{j=1}^{n} W_{ij}},  \\
&\text{ where }
\mathbb{I} =
\begin{cases} 
1 & \text{if } i \in C_U(u) \text{ and } j \in C_V(u), \\ 
  & \text{i.e. if $i$ and $j$ are in the same module} \\
0 & \text{otherwise} 
\end{cases}
\end{align*}

The calculation of $Q$ assumes that weights are 0 or 1, the absence or existence of a weight.
Since we are interested in a modularity metric for weighted graphs (weight matrices), 
we want to use a more `continuous' metric, and we want to punish large weights more heavily than small weights.

\end{document}